\documentclass{article}

\usepackage{arxiv}

\usepackage[utf8]{inputenc} 
\usepackage[T1]{fontenc}    
\usepackage{hyperref}       
\usepackage{url}            
\usepackage{booktabs}       
\usepackage{amsfonts}       
\usepackage{nicefrac}       
\usepackage{microtype}      
\usepackage{lipsum}		
\usepackage{graphicx}
\usepackage{natbib}
\usepackage{doi}
\usepackage{multirow}
\usepackage{subfigure}
\usepackage{amssymb}
\usepackage{amsmath}

\pdfoutput=1
\hypersetup{
    colorlinks=true,
    linkcolor=black,
    citecolor=black,
    filecolor=black,
    urlcolor=black,
}

\title{A Deep Learning Approach Based on Graphs to Detect Plantation Lines}

\author{ \href{https://orcid.org/0000-0002-4527-5724}{\includegraphics[scale=0.06]{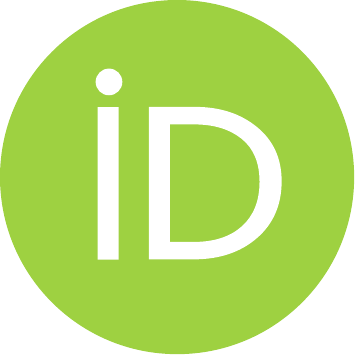}\hspace{1mm}Diogo Nunes Gon\c{c}alves}\\
	Federal University of Mato Grosso do Sul\\
	Campo Grande, MS, Brazil \\
	\And
	\href{https://orcid.org/0000-0002-2273-8968}{\includegraphics[scale=0.06]{orcid.pdf}\hspace{1mm}Mauro dos Santos de Arruda} \\
	Federal University of Mato Grosso do Sul\\
	Campo Grande, MS, Brazil \\
	\And
	\href{https://orcid.org/0000-0001-8181-760X}{\includegraphics[scale=0.06]{orcid.pdf}\hspace{1mm}Hemerson Pistori} \\
	Catholic University of Dom Bosco\\
	Campo Grande, MS, Brazil \\
	\And	
	\href{https://orcid.org/0000-0001-6633-2903}{\includegraphics[scale=0.06]{orcid.pdf}\hspace{1mm}Vanessa Jordão Marcato Fernandes} \\
	Federal University of Grande Dourados\\
	Dourados, MS, Brazil \\
	\And
	\href{https://orcid.org/0000-0001-6633-2903}{\includegraphics[scale=0.06]{orcid.pdf}\hspace{1mm}Ana Paula Marques Ramos} \\
	University of Western São Paulo\\
	Presidente Prudente, SP, Brazil \\
	\And
	\href{https://orcid.org/0000-0002-9106-1620}{\includegraphics[scale=0.06]{orcid.pdf}\hspace{1mm}Danielle Elis Garcia Furuya} \\
	University of Western São Paulo\\
	Presidente Prudente, SP, Brazil \\
	\And
    \href{https://orcid.org/0000-0002-0258-536X}{\includegraphics[scale=0.06]{orcid.pdf}\hspace{1mm}Lucas Prado Osco}\thanks{corresponding author: lucasosco@unoeste.br} \\
	University of Western São Paulo\\
	Presidente Prudente, SP, Brazil \\
	\And
	\href{https://orcid.org/0000-0000-0000-0000}{\includegraphics[scale=0.06]{orcid.pdf}\hspace{1mm}Hongjie He} \\
	University of Waterloo\\
	Waterloo, ON, Canada \\
	\And
	\href{https://orcid.org/0000-0001-7899-0049}{\includegraphics[scale=0.06]{orcid.pdf}\hspace{1mm}Jonathan Li} \\
	University of Waterloo\\
	Waterloo, ON, Canada \\
	\And
	\href{https://orcid.org/0000-0002-9096-6866}{\includegraphics[scale=0.06]{orcid.pdf}\hspace{1mm}José Marcato Junior} \\
	Federal University of Mato Grosso do Sul\\
	Campo Grande, MS, Brazil \\
	\And
	\href{https://orcid.org/0000-0002-8815-6653}{\includegraphics[scale=0.06]{orcid.pdf}\hspace{1mm}Wesley Nunes Gonçalves} \\
	Federal University of Mato Grosso do Sul\\
	Campo Grande, MS, Brazil \\
}

\renewcommand{\shorttitle}{\textit{arXiv - Gonçalves et al. (2021)}}

\hypersetup{
pdftitle={A Deep Learning Approach Based on Graphs to Detect Plantation Lines},
pdfsubject={cs.CV},
pdfauthor={Gonçalves, L.P. et al.},
pdfkeywords={remote sensing, convolutional neural networks, aerial imagery, UAV, object detection},
}

\begin{document}
\maketitle

\begin{abstract}
Identifying plantation lines in aerial images of agricultural landscapes is required for many automatic farming processes. Deep learning-based networks are among the most prominent methods to learn such patterns and extract this type of information from diverse imagery conditions. However, even state-of-the-art methods may stumble in complex plantation patterns. Here, we propose a deep learning approach based on graphs to detect plantation lines in UAV-based RGB imagery presenting a challenging scenario containing spaced plants. The first module of our method extracts a feature map throughout the backbone, which consists of the initial layers of the VGG16. This feature map is used as an input to the Knowledge Estimation Module (KEM), organized in three concatenated branches for detecting 1) the plant positions, 2) the plantation lines, and 3) for the displacement vectors between the plants. A graph modeling is applied considering each plant position on the image as vertices, and edges are formed between two vertices (i.e. plants). Finally, the edge is classified as pertaining to a certain plantation line based on three probabilities (higher than 0.5): i) in visual features obtained from the backbone; ii) a chance that the edge pixels belong to a line, from the KEM step; and iii) an alignment of the displacement vectors with the edge, also from KEM. Experiments were conducted in corn plantations with different growth stages and patterns with aerial RGB imagery. A total of 564 patches with 256 x 256 pixels were used and randomly divided into training, validation, and testing sets in a proportion of 60\%, 20\%, and 20\%, respectively. The proposed method was compared against state-of-the-art deep learning methods, and achieved superior performance with a significant margin, returning precision, recall, and F1-score of 98.7\%, 91.9\%, and 95.1\%, respectively. This approach is useful in extracting lines with spaced plantation patterns and could be implemented in scenarios where plantation gaps occur, generating lines with few-to-none interruptions.
\end{abstract}


\section{Introduction}

Linear objects, also denominated linear features in the photogrammetric context, are common in images, especially in anthropic scenes. Consequently, they are used in several photogrammetric tasks, such as orientation or triangulation \citep{KUBIK1991} \citep{SCHENK2004} \citep{Tommaselli2012} \citep{MARCATOJUNIOR2013} \citep{SUN2019} \citep{YAVARI2018}, rectification \citep{6709786} \citep{LONG2015}, matching \citep{WEI2021}, restitution \citep{LEE2004}, and camera calibration \citep{8340844} \citep{Badapour2017}. The registration of images and LiDAR (Light Detection And Ranging) data is also a topic that benefits from this type of object \citep{Habib2005} \citep{YANG2015}. Previous works proposed several approaches to automatically detect lines in images based on traditional digital image processing techniques. But, these approaches usually require a significant number of parameters and are not always robust when dealing with challenging situations, including shadows, pixel-pattern, geometry, among others. 

In recent years, artificial intelligence methods, especially deep learning-based algorithms, have been adapted to process remote sensing images from several spatial-spectral-resolution traits, aiming to attend distinct application areas. Deep learning, which is a subarea of machine learning, are state-of-the-art methods well-known for their ability to deal with challenging and varied tasks, involving scene-wise recognition, object detection, and semantic segmentation problems \citep{osco2021DL}. For each of these problems-domain specifics, several attempts have been made and great results were found. As such, deep neural networks (DNN) are quickly becoming one of the most prominent paths to learn and extract information from remote sensing data. This is mainly because it is difficult for the same method to evaluate different domains with the same performance, while deep learning developments aim to produce intelligent and robust mechanisms to deal with multiple learning patterns.

Based on recent literature analysis in photogrammetry and remote sensing fields few studies were developed focusing on deep learning for the detection of linear objects. Some deep networks based on segmentation approaches were proposed for line pattern detection, being mostly of them in road and watercourse extraction of aerial or orbital imagery. \citep{Yang2019} developed a multitask learning method to simultaneously segment roads and to detect its respective centerlines. Their framework was based on recurrent neural networks and the U-Net method \citep{Ronneberger2015}. A more recent publication \citep{Wei2020} proposed an innovative solution to segment and detect roads centerlines. Similarly, semantic segmentation approaches were developed in environmental applications with linear patterns, like river margin extraction in remote sensing imagery. An investigation \citep{Xia2019} proposed a deep network adopting the ResNet \citep{Resnet16} as the backbone of their framework for river segmentation in orbital images of medium-resolution. Another study \citep{Weng2020} presented a separable residual SegNet \citep{Badrinarayanan2017} method to segment rivers in remote sensing images, showing significant improvements over other deep learning-based approaches, including FCN \citep{Long_2015_CVPR} and DeconvNet \citep{Noh2015}. \citep{Wei2020b} developed a semantic distance-based segmentation approach to extract rivers in images obtaining an F1-score superior to 93\%, which outperformed several state-of-the-art algorithms.

In agricultural applications, a previous related-work \citep{osco2021cnn} proposed a method to simultaneously detect plants and plantation-lines in the agriculture field using UAV (Unmanned Aerial Vehicle) imagery datasets through deep learning algorithms. However, for this task, only visual features like texture, color, shape, etc. of the plants and plantation lines were considered by the DNN algorithm. Consequently, the plants’ locations (i.e. points) from different plantation lines were considered as belonging to the same line due to their proximity to one another. This, however, indicated a limited potential of this approach mainly when gaps or adverse patterns in the plantations occurred. In other agricultural-related remote sensing tasks, \citep{Rosa2020} proposed an approach based on semantic segmentation associated with geometric features to detect citrus plantation lines. Still, segmentation-based methods are not adequate to deal with spaced plants (non-continuous objects), which is the case of most crops in the initial stage. Moreover, another problem is when plantation gaps occur in later stages, wherein, for instance, plants are removed due to diseases or environmental hazards (e.g. strong winds). Additionally, line extraction, when associated with gap detection, is essential to conduct the replanting process, minimizing the losses in the cultivars, but that still remains an unsolved question inside both remote sensing and photogrammetric contexts supported by deep learning approaches.

A potential alternative that may support the aforementioned issues in regard to differences in patterns and space between the objects (e.g. plants, for instance) is the adoption of graph theory at the learning and extraction processes. Graphs are a type of structure that considers that some pairs of objects are related in a given feature or real-world space scene. Therefore, they can be useful for representing the relationship between objects in multiple domains, and can even inherit complicated structures containing rich underlying values \citep{Zhang2020}. As such, recent deep learning-based approaches have been proposed to evaluate or implement graph patterns for distinct problems-domain. Some of these approaches include strategies related to graph convolutional and/or recurrent neural networks, graph autoencoders, graph reinforcement learning, graph adversarial methods, and others \citep{Zhang2020}. Since graphs work by representing both the domain concept and their relationships, it makes them an innovative approach for improving the inference ability of objects in remote sensing imagery. Therefore, the combination of graph reasoning with the deep learning capability may work as complementary advantages of both techniques.

It is worth mentioning that few recent investigations integrating graphs in deep network models were conducted in the remote sensing and photogrammetry domain. One of which demonstrated the potential of implementing a semantic segmentation network with a graph convolutional neural network (CNN) to perform the segmentation of urban aerial imagery, identifying features like vegetation, pavement, buildings, water, vehicles, as others \citep{Ouyang2021}. Another study \citep{Ma2019} used an attention graph convolution network to segment land covers from SAR imagery, which demonstrated its high potential. A graph convolutional network was also used in a scene-wise classification task \citep{Gao2021}, discriminating between varied scenes from publicly available repositories containing images from several examples of land cover. In the hyperspectral domain, one approach \citep{Hong2020} was capable of successfully presenting a graph convolutional network-based method to pixel-wise classify differential land cover in urban environments. Furthermore, in urban areas, a  graph convolutional neural network was investigated to classify building patterns using spatial vector data \citep{Yan2019}. In the agricultural context, a cross-attention mechanism was adopted with a graph convolution network \citep{9210789} to separate (scene-wise classification) different crops, such as soybeans, corn, wheat, woods, hay, and others. The results were compared against state-of-the-art deep learning networks, outperforming them.

Regardless, up to the time of writing, no single-step approach to a DNN architecture was proposed with the integration of graphs to solve issues related to the identification and refinement of a plantation-line position in remote sensing imagery. In this paper, we propose a novel deep learning method based on graphs that estimate the displacement vectors linking one plant to another on the same plantation line. Three information branches were considered, being the first used for extracting the plants’ positions, the second for extracting the plantation lines, and the third for the displacement vectors. To demonstrate this approach effectiveness, experiments were conducted within a corn plantation field, at different growth stages, where some plantation-gaps were identified due to problems that occurred during the planting process. Moreover, to verify the robustness of our method with the addition of graphs, we compared it against both a baseline and other state-of-the-art deep neural networks, like \citep{lin2020deep} and \citep{Zhang_2019_CVPR}. Our study brings an innovative contribution related to extract plantation lines at challenging conditions, which may support several precision agriculture-related practices, since identifying plantation-lines in remote sensing images is a necessity for automatic farming processes.

The rest of this paper is organized as follows. In section 2, we detail the structure of our neural network and demonstrate how each step in its architecture is used in favor of extracting the plantation lines. In section 3, we present the results of the experiment, highlighting the performance of our  network in relation to its baselines, as well as comparing it against state-of-the-art deep learning-based methods. For section 4, we discuss in a broader tone the implications of implementing  graph information into our model, as well as indicating future perspectives in our approaches. Lastly, section 5 concludes the research presented here.

\section{Proposed Method}

\begin{figure}[ht]
 	\centering
 	\includegraphics[width=1.\columnwidth]{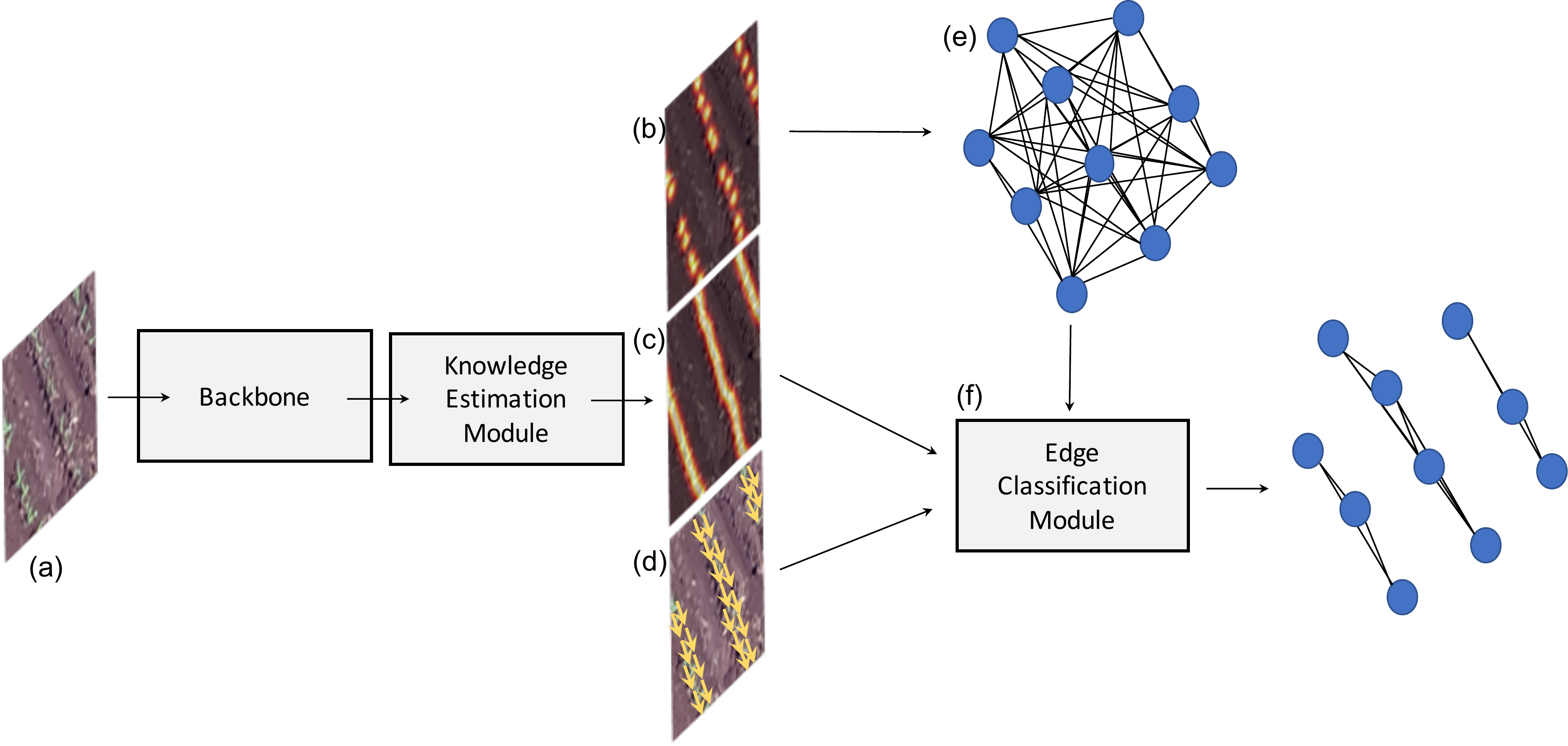}
 	\caption{The proposed approach composed of different modules.}
 	\label{fig:approach}
 \end{figure}

Initially, the proposed method estimates the necessary information from the input image using a backbone and a knowledge estimation module, as shown in Figure \ref{fig:approach}. The first information consists of a confidence map that corresponds to the probability of occurrence of plants in the image (Figure \ref{fig:approach} (b)). Through this confidence map, it is possible to estimate the position of each plant, which is useful in estimating the plantation lines. The second information corresponds to the probability that a pixel belongs to a crop line (Figure \ref{fig:approach} (c)). Finally, the third information is related to the estimate displacement of vectors linking one plant to another on the same plantation line (Figure \ref{fig:approach} (d)). These three information steps are relevant and help in detecting the plantation lines and estimating the number of plants in the image.

After these estimates, the problem of detecting plantation lines is modeled using a graph similar to \citep{Zhang_2019_CVPR}. Each plant identified in the confidence map is considered a vertex in the graph. The vertices/plants are connected forming a complete graph (Figure \ref{fig:approach} (e)). Each edge between two vertices is represented by a set of features extracted from the line that connects the two vertices in the image. These features and information from the knowledge estimation module are used in the edge classification module (Figure \ref{fig:approach} (f)) that classifies the edges as a planting line. The sections below describe these modules in detail. In Figure \ref{fig:approach}, the features are extracted from the image by means of a backbone and used to extract knowledge related to the position of each plant and line, in addition to displacement vectors between the plants. The position of each plant is modeled on a complete graph and each edge is classified based on the extracted knowledge.

\subsection{Backbone - Feature Map Extraction}

\begin{figure}[htb]
	\centering
	\includegraphics[width=1.\columnwidth]{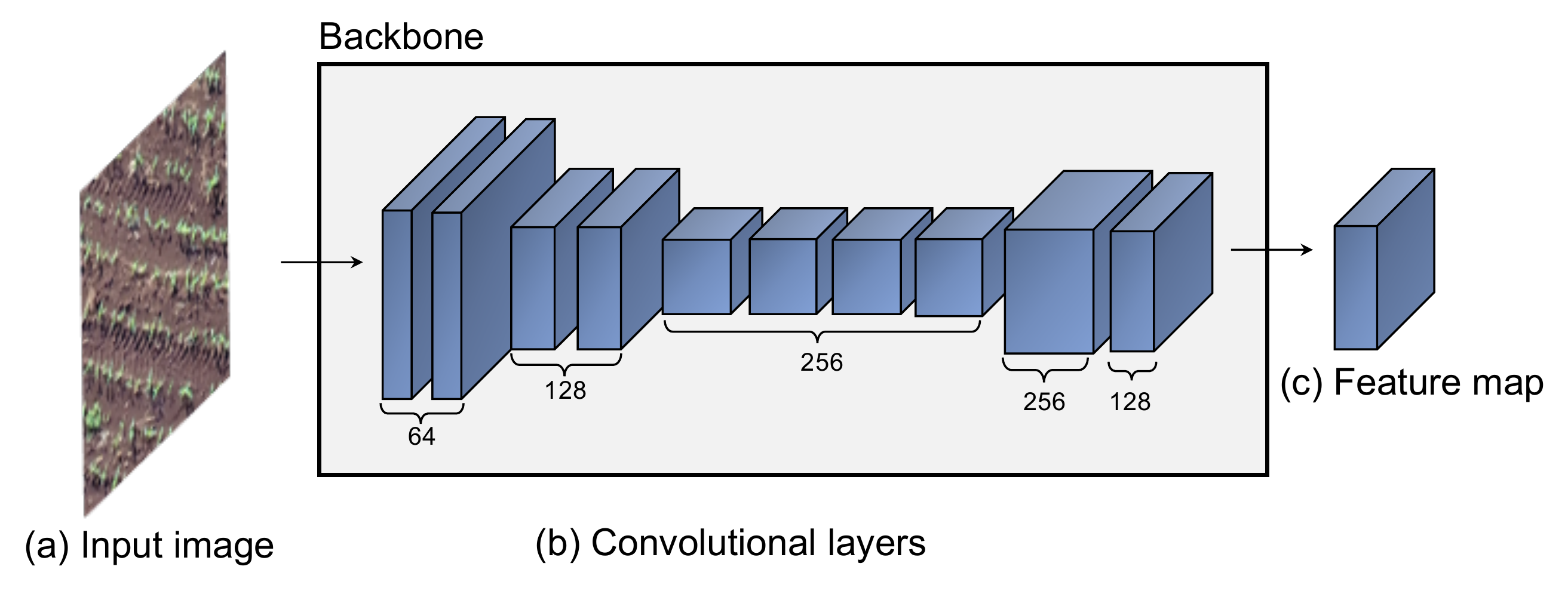}
	\caption{The backbone used in the proposed structure is composed of the initial layers of VGG16 and a bilinear upsampling layer.}
	\label{fig:backbone}
\end{figure}

The first module of the proposed method consists of extracting a feature map $F$ through a backbone as shown in Figure \ref{fig:backbone}. In this work, the backbone consists of the initial layers of the VGG16 network \citep{VGG15}. The first and second convolutional layers have 64 filters of size $3 \times 3$ and are followed by a max-pooling layer with window $2 \times 2$. Similarly, six convolutional layers (two with 128 filters and four with 256 filters of size $3 \times 3$) and a max-pooling layer are then applied. To obtain a resolution large enough, a bilinear upsampling layer is applied to double the resolution of the feature map. Finally, two convolutional layers with 256 and 128 $3 \times 3$ filters are used to obtain a feature map that describes the image content. All convolutional layers have the ReLU activation function (Rectified Linear Units). Given an input image $I$ with resolution $w \times h$, a feature map $F$ with resolution $\frac{w}{2} \times \frac{h}{2}$ is obtained.

\subsection{Knowledge Estimation Module (KEM)}

\begin{figure}[ht]
	\centering
	\includegraphics[width=1.\columnwidth]{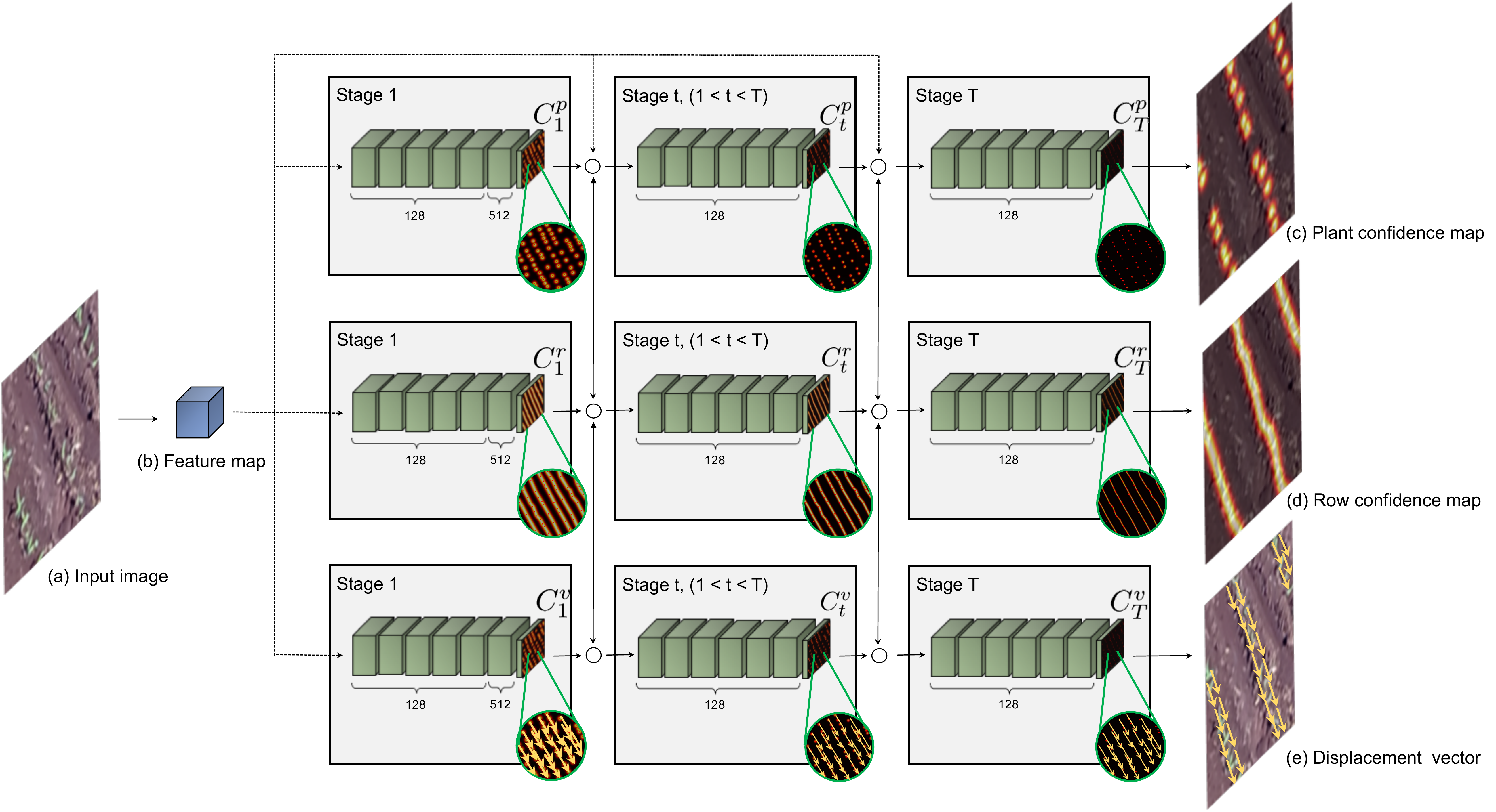}
	\caption{Knowledge estimation module related to a confidence map of plants and plantation lines, in addition to displacement vectors between plants on the same line.}
	\label{fig:kem}
\end{figure}

The feature map $F$ is used as an input to the Knowledge Estimation Module - KEM (Figure \ref{fig:kem}). The information is estimated through three branches, each branch consisting of $T$ stages. The first stage of each branch receives the feature map $F$ and estimates a confidence map for the plant positions $C_1^{p}$ (first branch), a confidence map for plantation lines $C_1^{r}$ (second branch), and the displacement vectors $C_1^{v}$ that connect a plant to another on the same plantation line (third branch). The estimation in the first stage is performed by five convolutional layers: three layers with 128 filters of size $3 \times 3$ and one layer with 512 filters of size $1 \times 1$. The $1 \times 1$ filter can perform a channel-wise information fusion and dimensionality reduction to save computational cost. Finally, the last layer has a single filter for estimating plants $C_1^{p}$ and plantation lines $C_1^{r}$, and two filters (i.e., displacement in $x,y$) for the displacement vectors $C_1^{v}$.

At a later stage $t$, the estimates from the previous stage $C_{t-1}^{p}, C_{t-1}^{r}, C_{t-1}^{v}$ and the feature map $F$ are concatenated and used to refine the estimates $C_{t}^{p}, C_{t}^{r}, C_{t}^{v}$. The $T-1$ final stages consist of seven convolutional layers, five layers with 128 filters of size $7 \times 7$, one layer with 128 filters of size $1 \times 1$ and the final layer for estimation according to the first stage.

The multiple stages assist in hierarchical and collaborative learning in estimating the occurrence of plants, lines and displacement vectors \citep{Osco2020, osco2021cnn}. The first stage performs the rough prediction of the information that is refined in the later stages.

\subsection{Graph Modeling}

The problem of detecting plantation lines is modeled by a graph $G = (V, E)$ composed of a set of vertices $V =  \{v_i \}$ and edges $E = \{e_{ij} \}$. Each detected plant is represented by a vertex $v_i = (x_i, y_i)$ with the spatial position of the plant in the image. The vertices are connected to each other forming a complete graph.

The plants are obtained from the confidence map of the last stage, $C_{T}^{p}$. For this, the peaks (local maximum) are estimated from $C_{T}^{p}$ by analyzing a 4-pixel neighborhood. Thus, a pixel is a local maximum if $C_{T}^p(x,y) > C_{T}^p(l,m)$ for all neighbors given by $(l \pm 1, m)$ or $(l, m \pm 1)$. To avoid detecting plants with a low probability of occurrence, a plant is detected only if $C_{T}^p(x,y) > \tau$. In addition, we consider a minimum distance $\delta$ so that the detection of very close plants does not occur. After preliminary experiments, we set $\tau = 0.15$ and $\delta = 1$ pixels.

\subsection{Edge Classification Module (ECM)}

\begin{figure}[ht]
	\centering
	\includegraphics[width=1.\columnwidth]{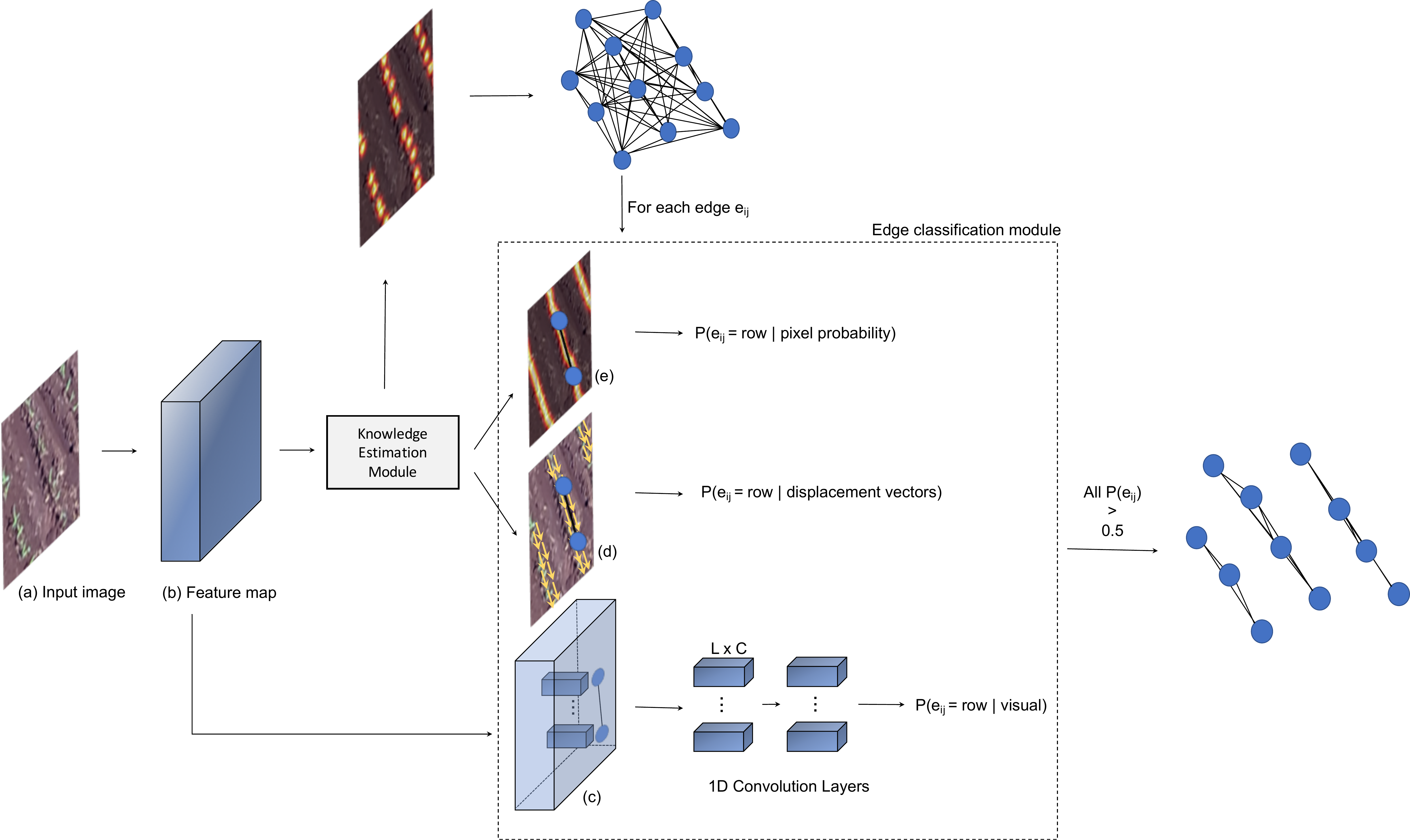}
	\caption{Module for extracting features and classifying an edge $e_{ij}$.}
	\label{fig:fwm}
\end{figure}

Given the complete graph, the detection of plantation lines consists of classifying each edge (Figure \ref{fig:fwm}). Here, the feature vectors of the backbone are sampled from the line connecting the vertices $i$ and $j$  and from the estimates made by the knowledge estimation module. This information is used to classify an edge as a plantation line. Each edge $e_{ij}$ is equal to one (existing) only if the vertices $v_i$ and $v_j$ (i.e., plants $i$ and $j$) belong to the same plantation line. For this, this module estimates three probabilities of a given edge belonging to a plantation line, being related to: i) visual features obtained from the backbone, ii) chance that the edge pixels belong to a line, and iii) alignment of the displacement vectors with the edge, the last two obtained by the knowledge estimation module. Therefore, an edge is classified as a plantation line if the three probabilities are greater than 0.5. The use of different characteristics for the edge classification makes it more robust. The subsections below describe the calculation of the three probabilities.

\subsubsection{Visual Features Probability}

Given an edge $e_{ij}$, $L$ equidistant points are sampled between $v_i = (x_i, y_i)$ and $v_j = (x_j, y_j)$. For each sampled point, a feature vector is obtained from the backbone activation map. In this way, each edge $e_{i,j}$ is represented by a set of features $F^{e_{i,j}} = \{ f_{1}^{e_{i,j}}, \dots, f_{l}^{e_{i,j}}, \dots, f_{L}^{e_{i,j}} \} \mid f_l^{e_{i,j}} \in \Re^{C}$, where $C$ is the number of channels in the activation map ($C=128$ in this work). To classify an edge using visual features, $F^{e_{ij}}$ is given as input for three 1D convolutional layers with $128, 256, 512$ filters. At the end, a fully connected layer with sigmoid activation corresponds to the probability of the edge belonging to a plantation line. Figure \ref{fig:fwm} (c) illustrates the process and the features that represent an edge.

\subsubsection{Displacement Vector Probability}

For each sampled point $l$ on edge $e_{ij}$, we measure the alignment between the line connecting $v_i$ and $v_j$ and the displacement vector at $l$. For the two vertices $v_i$ and $v_j$ of $e_{ij}$, we sample the displacement vectors predicted in $C_{T}^v$ along the line to calculate an association weight \citep{Cao2017}:

\begin{equation}
	\omega_l^{e_{ij}} = C_{T}^v(l) \cdot \frac{v_j - v_i}{\lVert v_j-v_i \rVert_2},
	\label{eq:vetor}
\end{equation}

where $C_{T}^v(l)$ corresponds to the displacement vector for the sampled point $l$ between $v_i$ and $v_j$. Finally, the edge probability based on the displacement vectors is given by the mean, $p(e_{ij} \mid \text{displacement vectors}) = \frac{1}{L} \sum_{l=1}^L \omega_l^{e_{ij}}$.

\begin{figure}[ht]
	\centering
	\subfigure[]{\label{fig:weightsa}\includegraphics[width=.49\columnwidth]{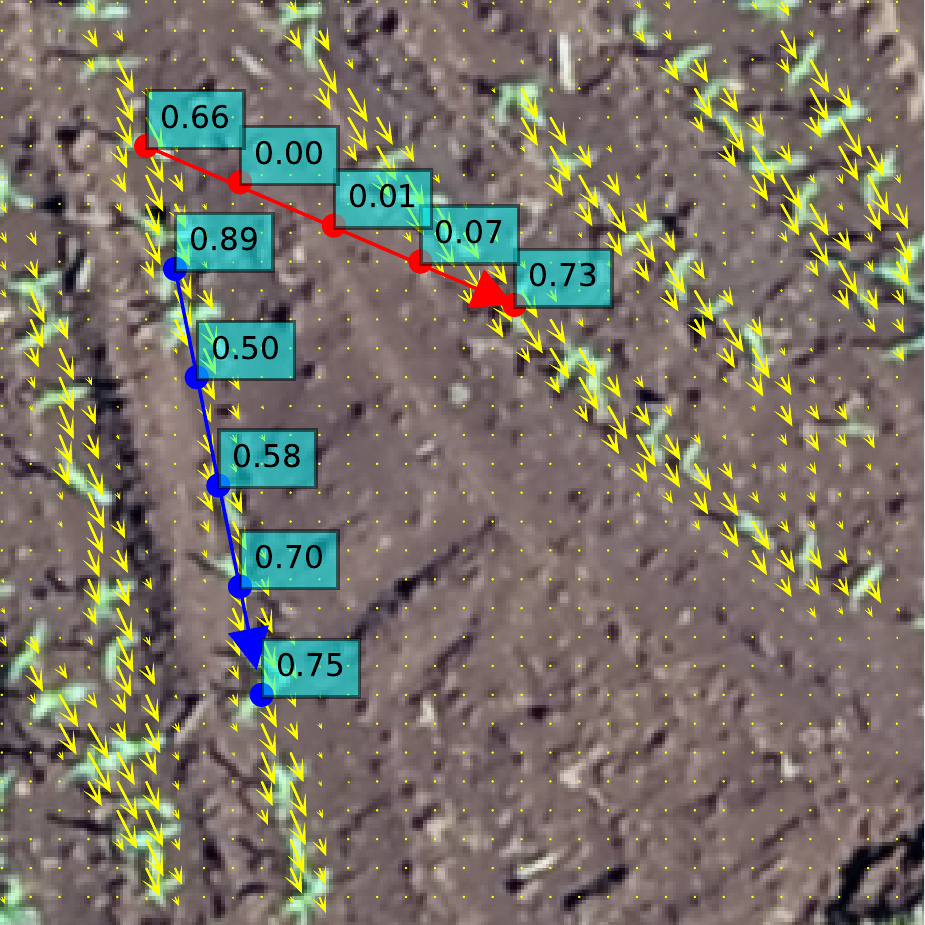}}
	\subfigure[]{\label{fig:weightsb}\includegraphics[width=.49\columnwidth]{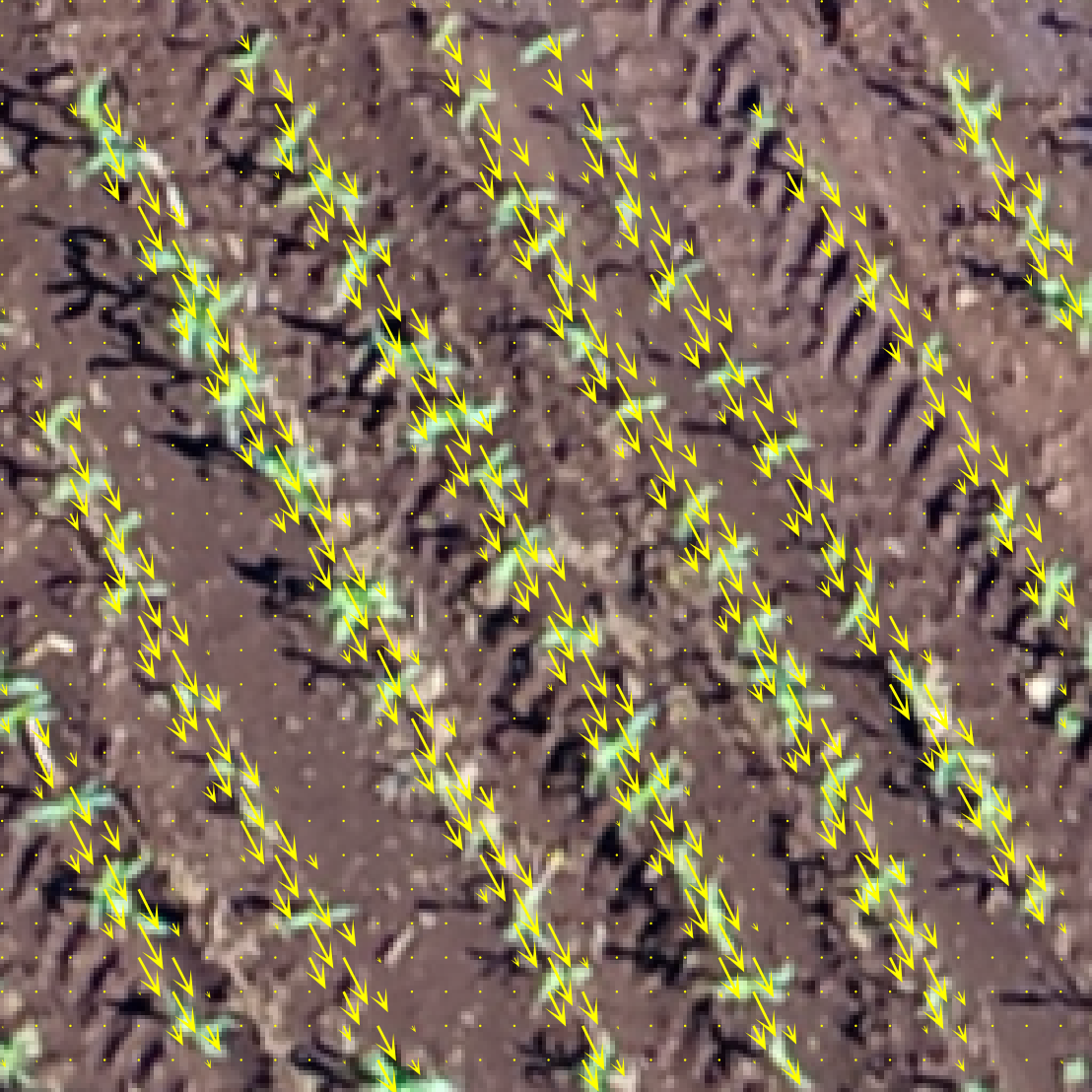}}
	\caption{(a) Example of the probability of two edges based on the displacement vectors and (b) example of the vectors estimated by the proposed method in a test image.}
\end{figure}

Figure \ref{fig:weightsa} illustrates the process for estimating the probability of an edge based on the displacement vectors. The blue edge connects two vertices/plants of the same plantation line while the red edge connects two vertices of different lines. For each edge, points are sampled along the line and the weights of the predicted vector alignment and the line connecting the vertices are shown. We can observe that points sampled in a plantation line tend to have a greater weight than points sampled in the background regions. As an illustration, Figure \ref{fig:weightsb} presents an example of the displacement vectors estimated by KEM for another test image.

\subsection{Pixel Probability}

This probability is calculated to estimate the edge importance based on the probability that the pixels are from a plantation line. Similarly to the previous section, we sample the points along the line $v_i$ and $v_j$ on the confidence map $C_{T}^r$ obtained from the KEM. The probability is given by the average of each sampled pixel $l$:

\begin{equation}
	p(e_{ij} \mid \text{pixel probability}) = \frac{1}{L} \sum_l C_{T}^r(l).
\end{equation}

Figure \ref{fig:weightsb} presents the calculation for two edges. We can see that the probability of a pixel belonging to a plantation line presents a good initial estimate, although it is not enough to obtain completely connected lines.

\begin{figure}[ht]
	\centering
	\includegraphics[width=.48\columnwidth]{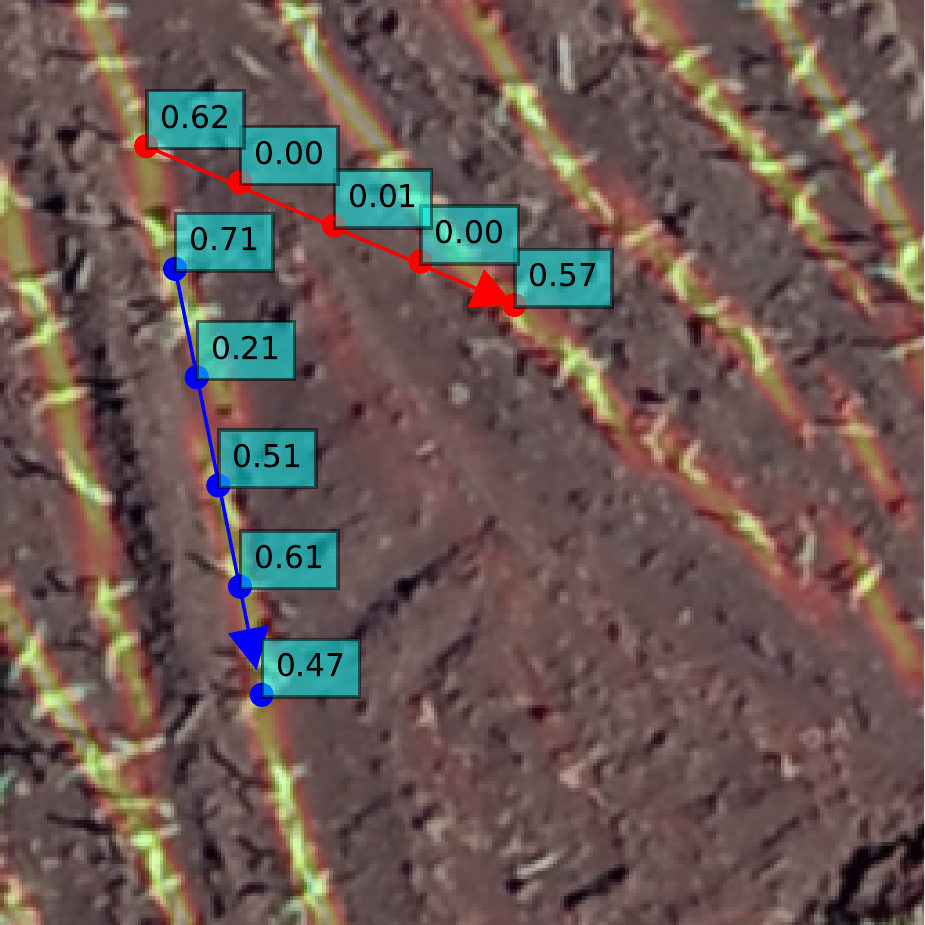}
	\caption{Calculation of the probability based on the chance of a pixel belonging to a plantation line.}
	\label{fig:weightsb}
\end{figure}

\subsection{Proposed Method Training}

Although the entire method can be trained end-to-end, we initially trained the knowledge estimation module (KEM). Next, we keep the KEM weights frozen and train the 1D convolutional layers of the edge classification module. This step-by-step training process was adopted to save computational resources. To train KEM, the loss function is applied at the end of each stage according to Equations \ref{eq:loss1}, \ref{eq:loss2} and \ref{eq:loss3} for the estimate made for the confidence map of the plant positions, line and displacement vectors, respectively. The overall loss function is given by Equation \ref{eq:loss4}.

\begin{align} 
	f_t^p = \sum_i \parallel \hat{C_t^p}(i) - C_{t}^p(i)\parallel^2_2
	\label{eq:loss1} \\
	f_t^r = \sum_i \parallel \hat{C_t^r}(i) - C_{t}^r(i)\parallel^2_2
	\label{eq:loss2} \\
	f_t^v = \sum_i \parallel \hat{C_t^v}(i) - C_{t}^v(i)\parallel^2_2
	\label{eq:loss3} \\
	f = \sum_{t=1}^{T} f_t^p + f_t^r + f_t^v
	\label{eq:loss4}
\end{align}
where $\hat{C_t}^p, \hat{C_t}^r$ and $\hat{C_t}^v$ are the ground truths for plant position, lines and displacement vectors, respectively.

$\hat{C_t}^p$ is generated for each stage $t$ by placing a Gaussian kernel in each center of the plants \citep{osco2021cnn}. The Gaussian kernel of each stage $t$ is different and has a standard deviation $\sigma_t$ equally spaced between $[\sigma_{max}, \sigma_{min}]$. In preliminary experiments, we defined $\sigma_{max} = 3$ and $\sigma_{min} = 1$. Similarly, $\hat{C_t}^r$ is generated considering all the pixels of the plantation lines and placing a Gaussian kernel with the same parameters as before. On the other hand, $\hat{C_t}^v$ is constructed using unit vectors. Given the position of two plants $v_i$ and $v_j$, the value $\hat{C_t}^v(l)$ of a pixel $l$ is a unit vector that points from $v_i$ to $v_j$ if $l$ lies on the line between the two plants and both belong to the same plantation line; otherwise, the value $\hat{C_t}^v(l)$ is a null vector. In practice, the set of pixels on the line between two plants is defined as those within the distance limit of the line segment (two pixels in this work).

\begin{figure}
	\centering
	\subfigure[]{\label{fig:gtsa}\includegraphics[width=.3\columnwidth]{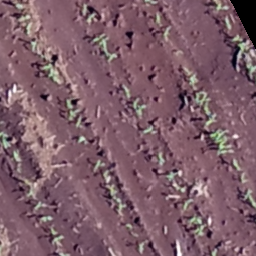}}\\
	\subfigure{\includegraphics[width=.3\columnwidth]{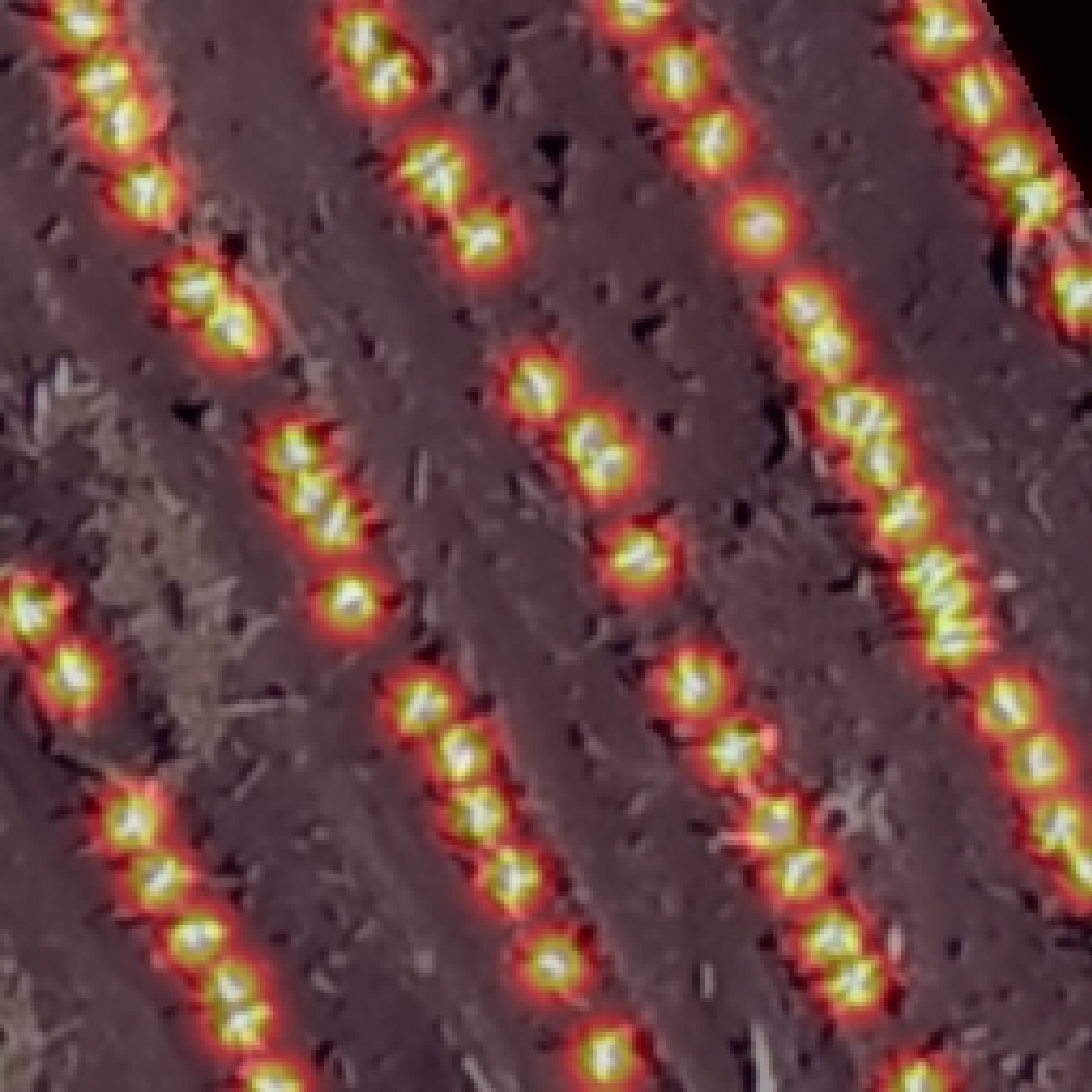}}
	\setcounter{subfigure}{1}
	\subfigure[]{\label{fig:gtsb}\includegraphics[width=.3\columnwidth]{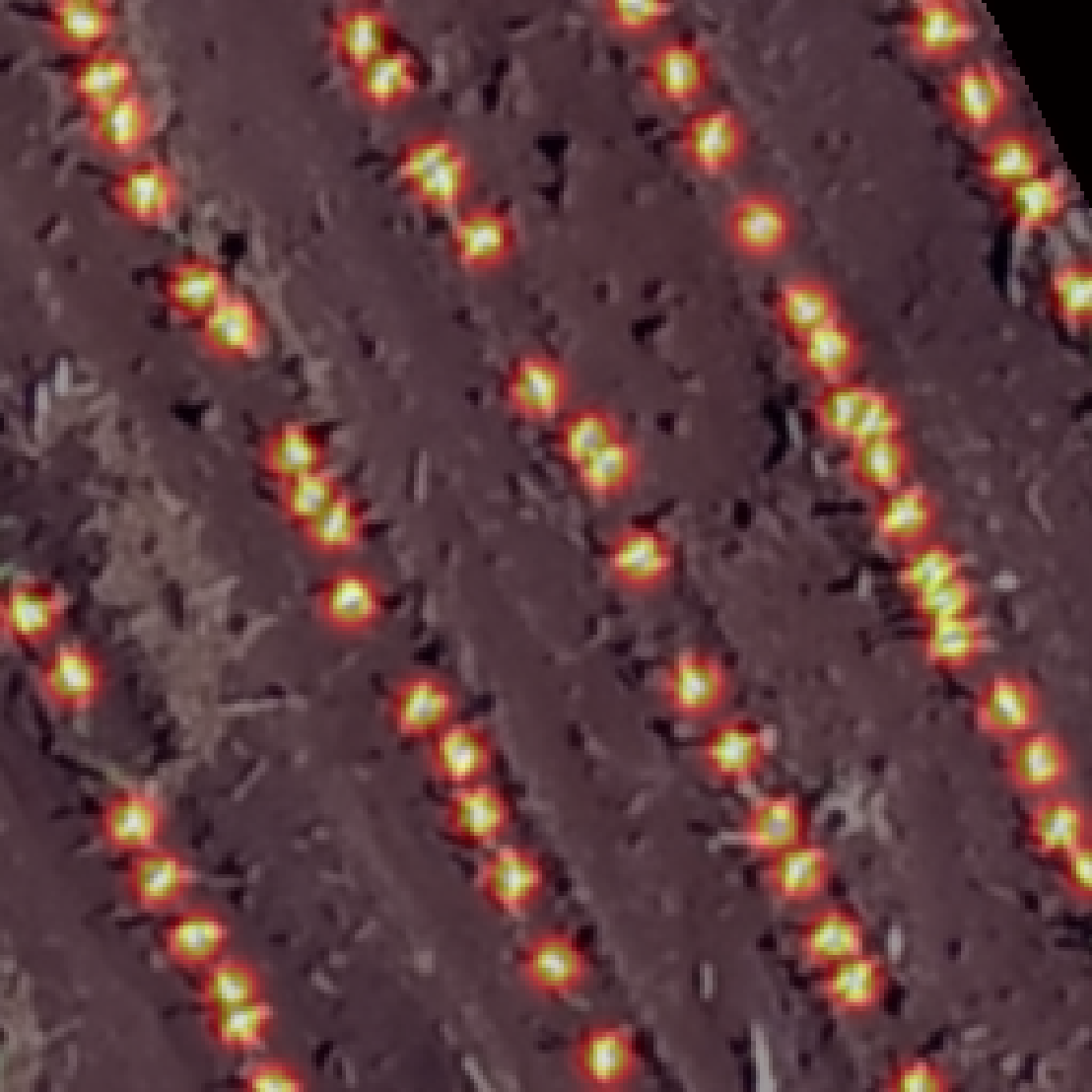}}
	\subfigure{\includegraphics[width=.3\columnwidth]{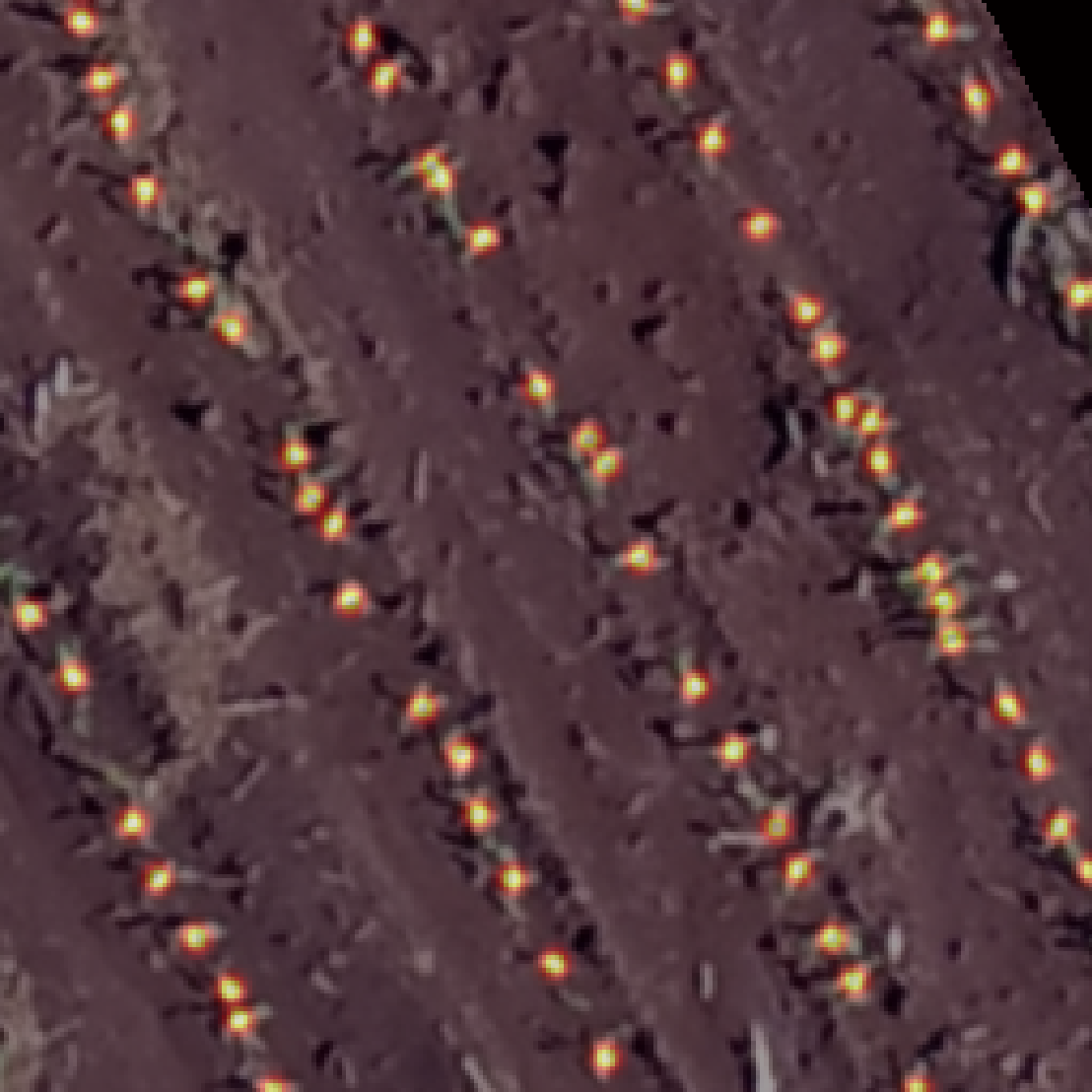}}
	\subfigure{\includegraphics[width=.3\columnwidth]{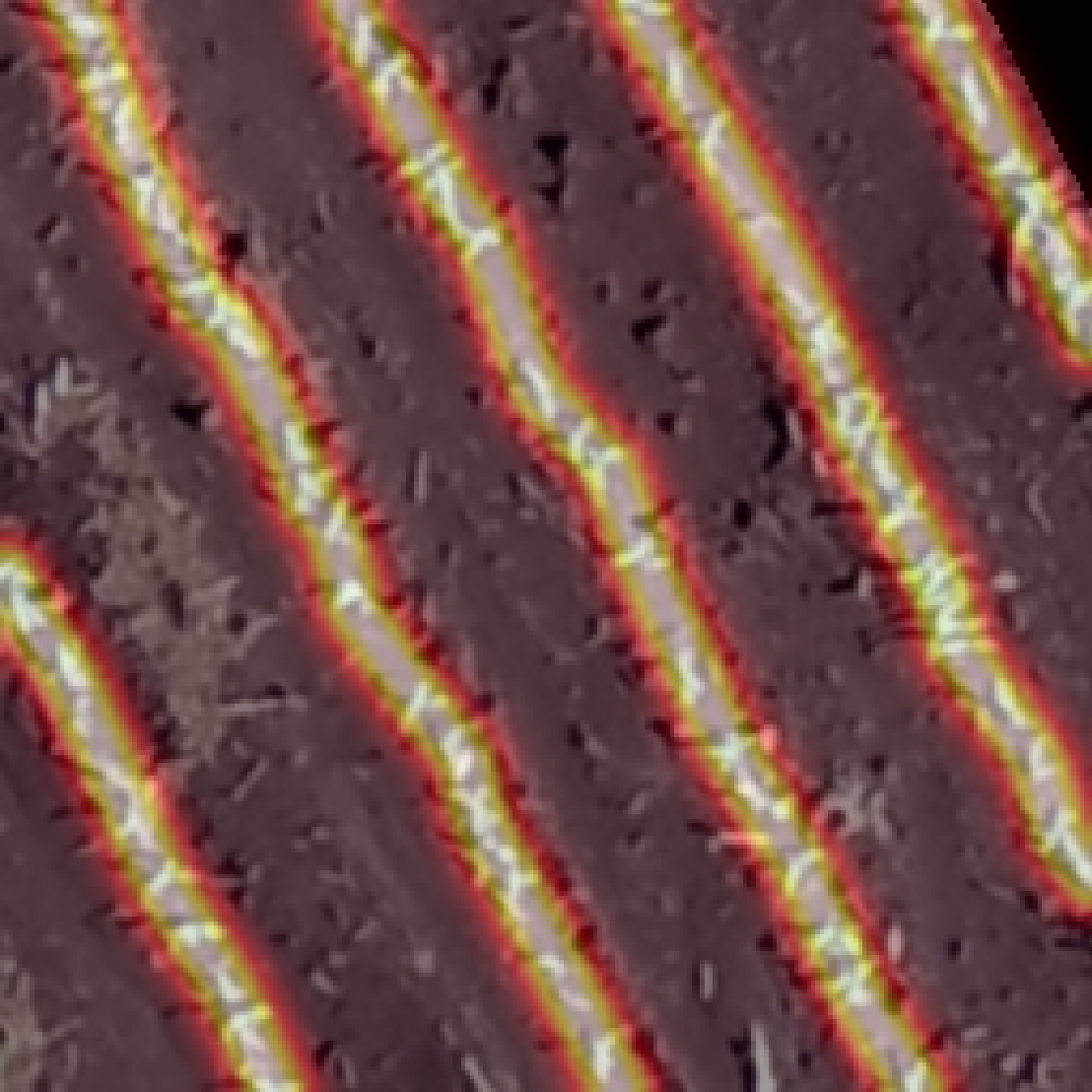}}
	\setcounter{subfigure}{2}
	\subfigure[]{\label{fig:gtsc}\includegraphics[width=.3\columnwidth]{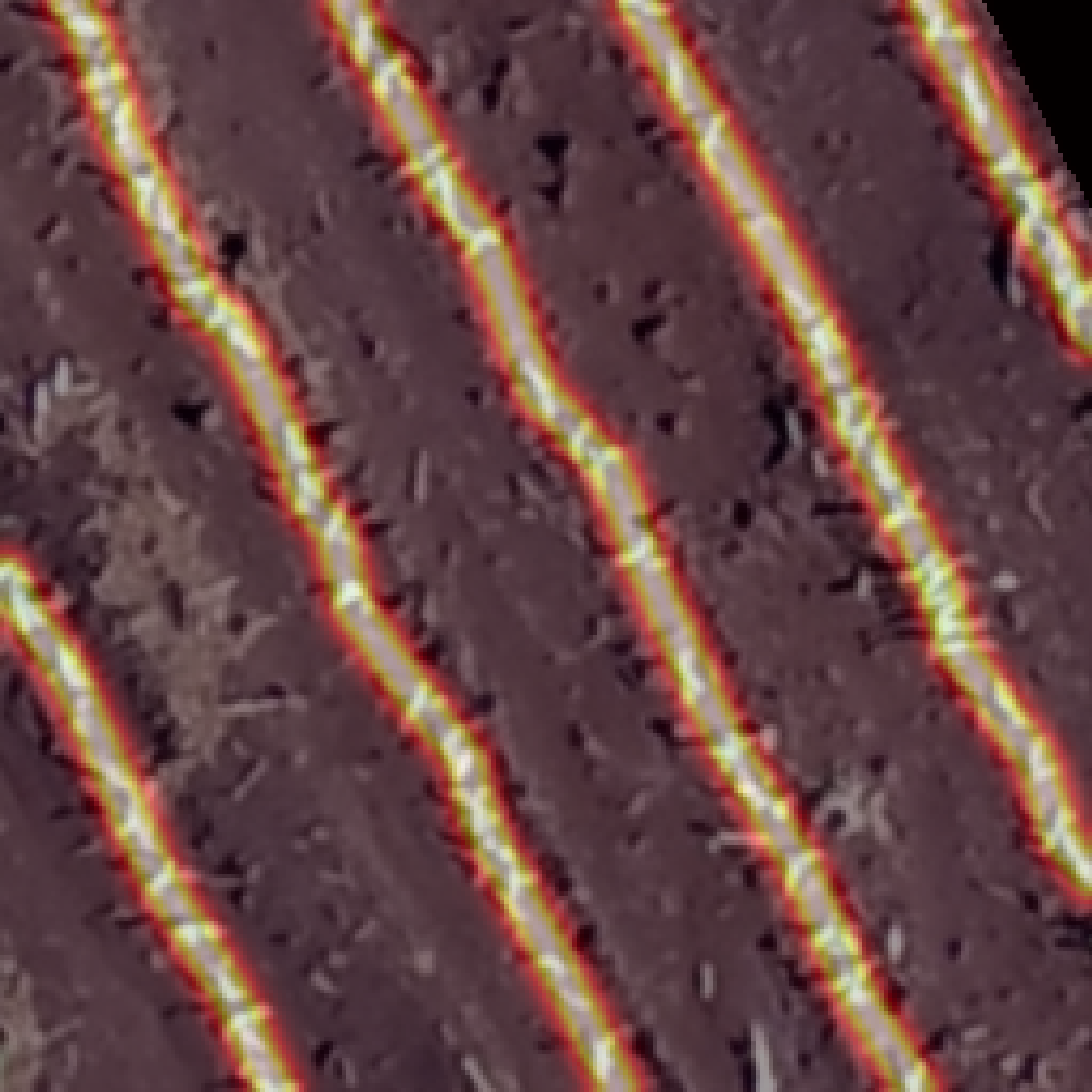}}
	\subfigure{\includegraphics[width=.3\columnwidth]{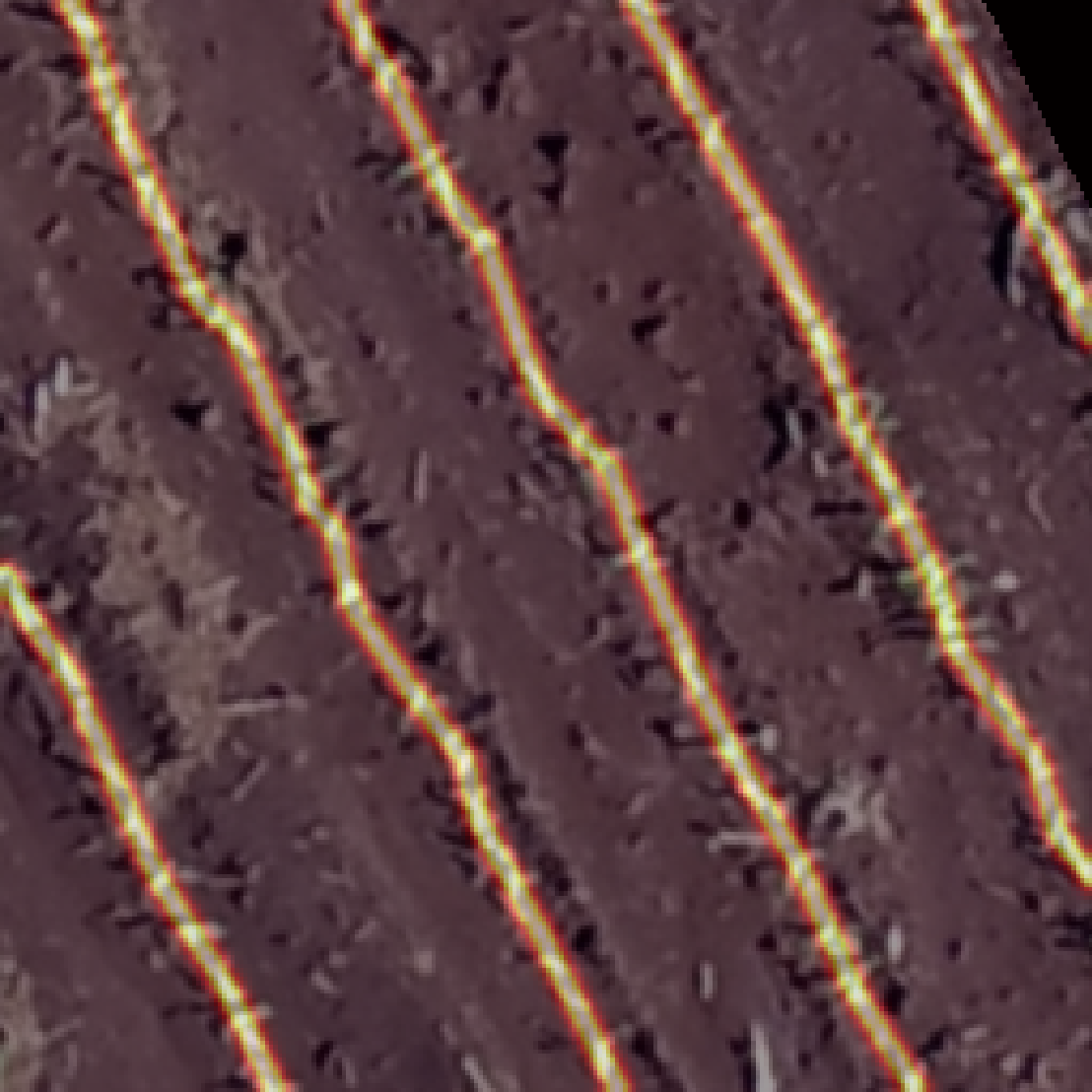}}
	\subfigure{\includegraphics[width=.3\columnwidth]{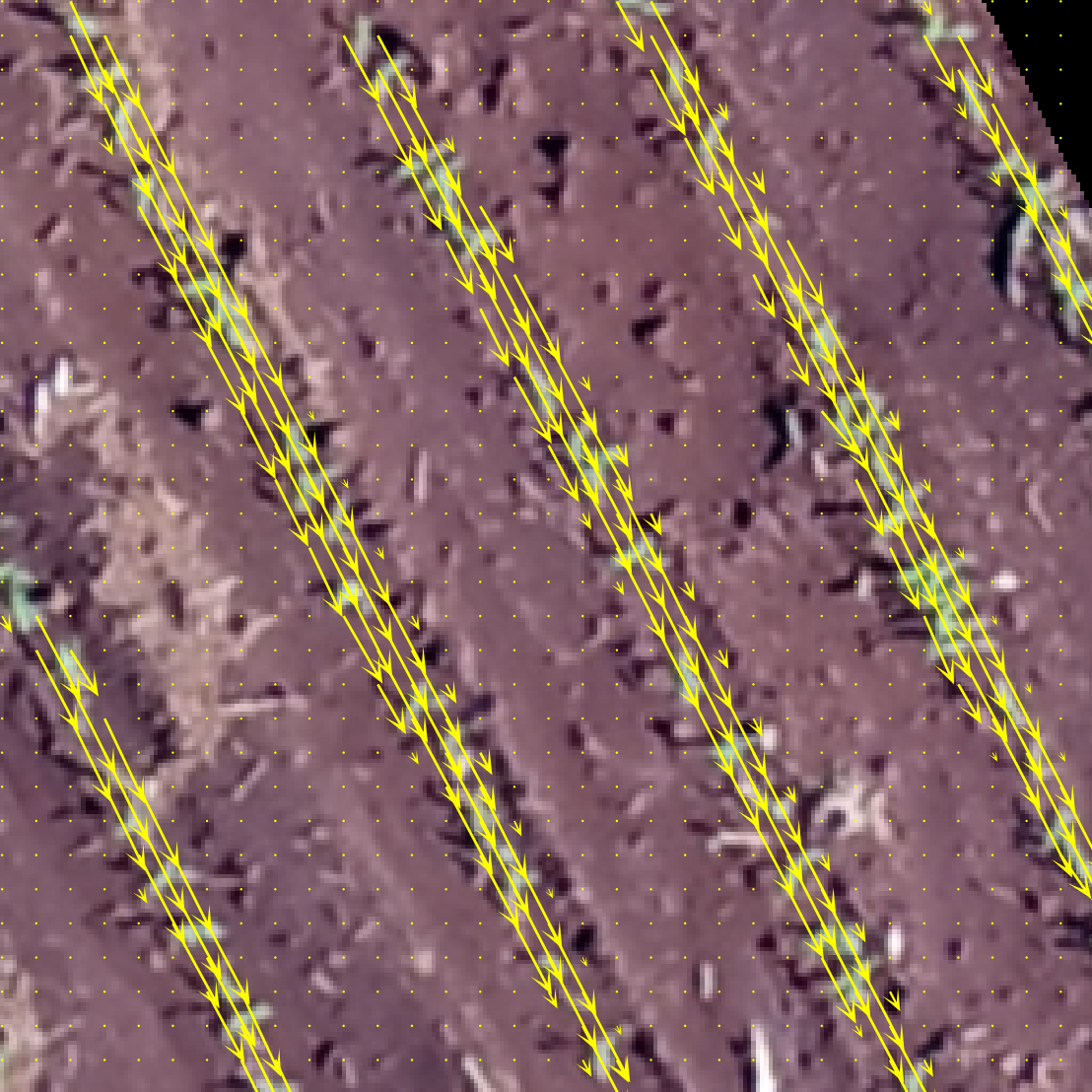}}
	\setcounter{subfigure}{3}
	\subfigure[]{\label{fig:gtsd}\includegraphics[width=.3\columnwidth]{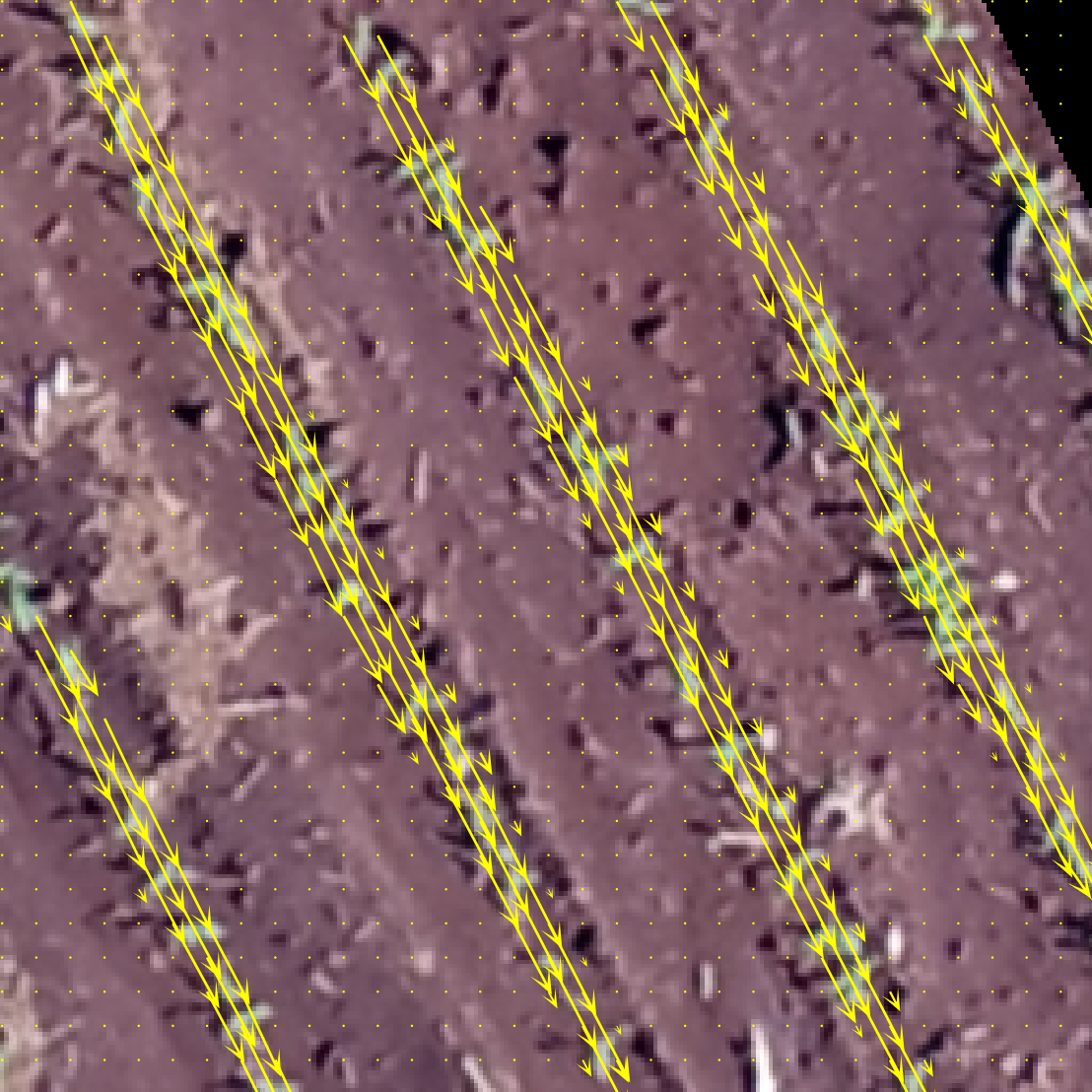}}
	\subfigure{\includegraphics[width=.3\columnwidth]{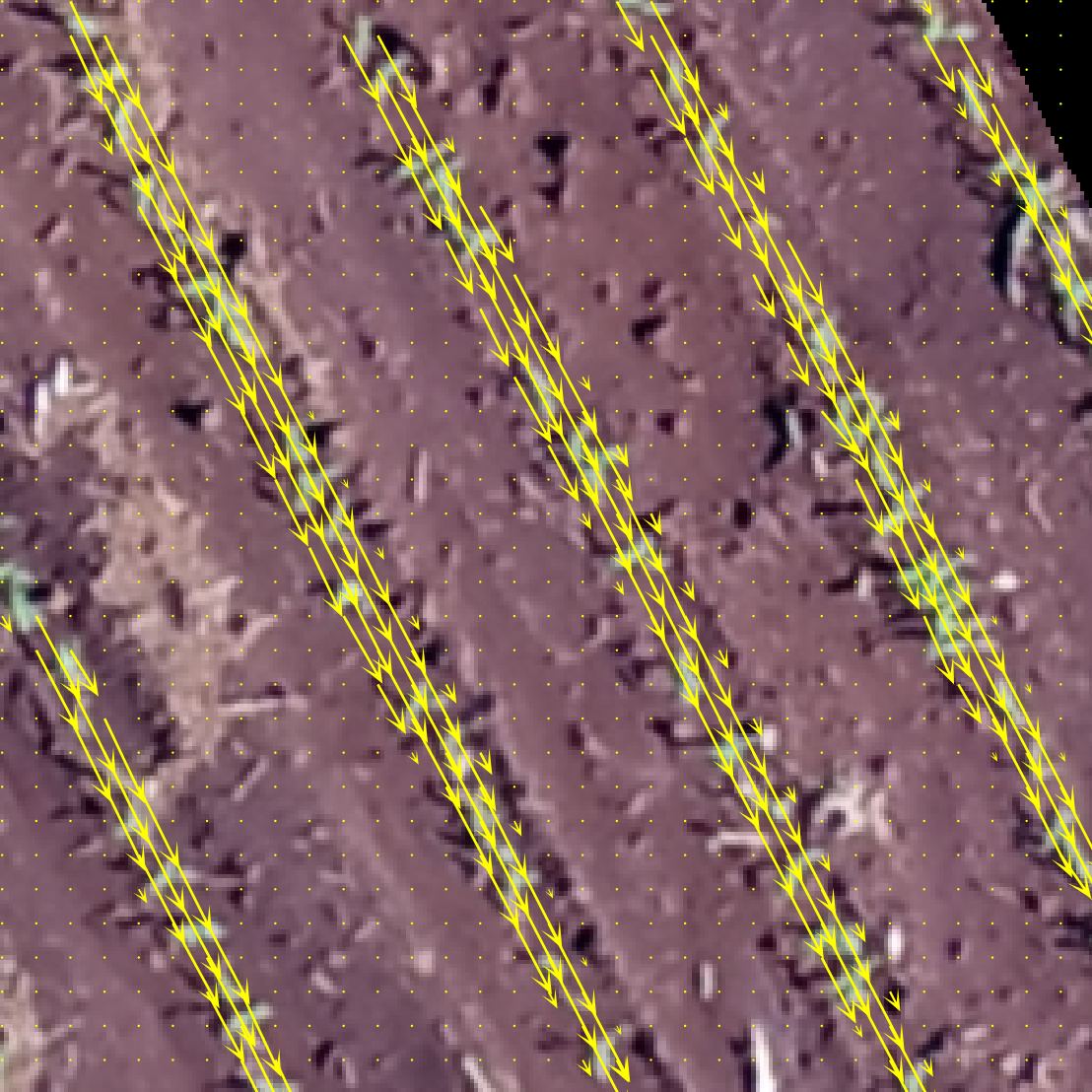}}
	\caption{(a) RGB image and ground truths for the three branches ((b) plant positions, (c) lines, (d) displacement vectors) and stages using different values for $\sigma$.}
	\label{fig:gts}
\end{figure}

Figure \ref{fig:gts} shows examples of ground truths for the three branches of KEM. The RGB image is shown in Figure \ref{fig:gtsa} while the ground truths for the branches and with three stages are shown in Figures \ref{fig:gtsb}, \ref{fig:gtsc} and \ref{fig:gtsd}.

The training of the 1D convolutional layers of the edge classification module is performed using binary cross-entropy loss. Given a set of features that describes an edge $e_{ij}$, its prediction $y_{e_{ij}}$ is obtained and compared with the ground truth $\hat{y}_{e_{ij}}$ (edge belongs or not to a plantation line) according to:

\begin{equation}
	loss = \hat{y}_{e_{ij}} \cdot \log y_{e_{ij}} + (1-\hat{y}_{e_{ij}}) \cdot \log (1-y_{e_{ij}}).
\end{equation}

\section{Experiments and Results}

\subsection{Experimental Setup}

\textbf{Image dataset:}
The image dataset used in the experiments was obtained from a previous work \citep{osco2021cnn}. The images were captured in an experimental area at “Fazenda Escola” at the Federal University of Mato Grosso do Sul, in Campo Grande, MS, Brazil. This area has approximately 7,435 m$^2$, with corn (Zea mays L.) plants planted at a $30 \times 50$ cm spacing, which results in 4-to-5 plants per square meter. For two days, the images were captured with a Phantom 4 Advanced (ADV) UAV using an RGB camera equipped with a 1-inch 20-megapixel CMOS sensor and processed with Pix4D commercial software. The UAV flight was approved by the Department of Airspace Control (DECEA) responsible for Brazilian airspace. The images were labeled by an expert initially detecting the plantation-lines. Then, each line was inspected and the plants were manually identified. The entire labeling process was carried out in the QGIS 3.10 open-source software.

The images were split into 564 patches with $256 \times 256$ pixels without overlapping. The patches were randomly divided into training, validation, and test sets, containing 60\%, 20\%, and 20\%, respectively. Since the patches have no overlap, it is guaranteed that no part of the images is repeated in different sets.

\textbf{Training:} 
The backbone weights were initialized with the VGG16 weights pretrained on ImageNet and all other weights were started at random. The methods were trained using stochastic gradient descent with a learning rate of 0.001, momentum of 0.9, and batch size of 4. KEM was trained using 100 epochs while the 1D convolutional layers of ECM was trained using 50 epochs. These parameters were defined after preliminary experiments with the validation set. The method was implemented in Python with the Keras-TensorFlow API. The experiments were performed on a computer with Intel (R) Xeon (E) E3-1270@3.80GHz CPU, 64 GB memory, and an NVIDIA Titan V graphics card, that includes 5120 CUDA (Compute Unified Device Architecture) cores and 12 GB of graphics memory.

\textbf{Metrics:}
To assess plant detection, we use the Mean Absolute Error (MAE), Precision, Recall and F1 (F-measure) commonly applied in the literature. These metrics can be calculated according to Equations \ref{eq:mae}, \ref{eq:precision}, \ref{eq:recall}, and \ref{eq:f1}.

\begin{equation}
	MAE = \frac{1}{N} \sum_i \mid n_i - m_i\mid
	\label{eq:mae}
\end{equation}
\begin{equation}
	Precision = \frac{TP}{TP+FP}
	\label{eq:precision}
\end{equation}
\begin{equation}
	Recall = \frac{TP}{TP+FN}
	\label{eq:recall}
\end{equation}
\begin{equation}
	F1 = 2 \cdot \frac{Precision \cdot Recall}{Precision + Recall}
	\label{eq:f1}
\end{equation}

where $N$ is the number of patches, $n_i$ is the number of plants labeled for patch $i$ and $m_i$ is the number of plants detected by a method. To calculate precision, recall and therefore F1, we need to calculate True Positive (TP), False Positive (FP), and False Negative (FN). For plant detection, TP corresponds to the number of plants correctly detected, while FP corresponds to the number of detections that are not plants and FN corresponds to the number of plants that were not detected by the method. A detected plant is correctly assigned to a labeled plant if the distance between them is less than $8$ pixels. This distance was estimated based on the plant canopy (see Figure \ref{fig:detecmilho} for examples).

Similarly, we use the Precision, Recall and F1 metrics to assess the detection of plantation lines. In contrast, the values of TP, FP and FN correspond to the number of pixels in a plantation line that have been correctly or incorrectly detected by the method compared to the labeled lines. A plantation line pixel is correctly assigned to a labeled one if the distance is less than $5$ pixels.

\subsection{Ablation Study}

In this section we individually evaluate the main modules of the proposed method. The first module is the plant detection that has a direct result in the construction of the graph. The next module consists of the edge classification and, in this step, the appropriate number of sampling points $L$ and the influence of each knowledge learned by KEM were evaluated.

\subsubsection{Plant Detection}

An important step of the proposed method is to detect the plants in the image that will compose the graph for later detection of the plantation lines. Detections occur by estimating the confidence map and detecting its peaks. The results of plant detection varying the number of KEM stages are shown in Table  \ref{tab:estagiosmilho}.

\begin{figure}[ht]
	\centering
	\subfigure{\includegraphics[width=.3\columnwidth]{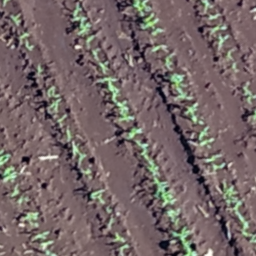}}
	\subfigure{\includegraphics[width=.3\columnwidth]{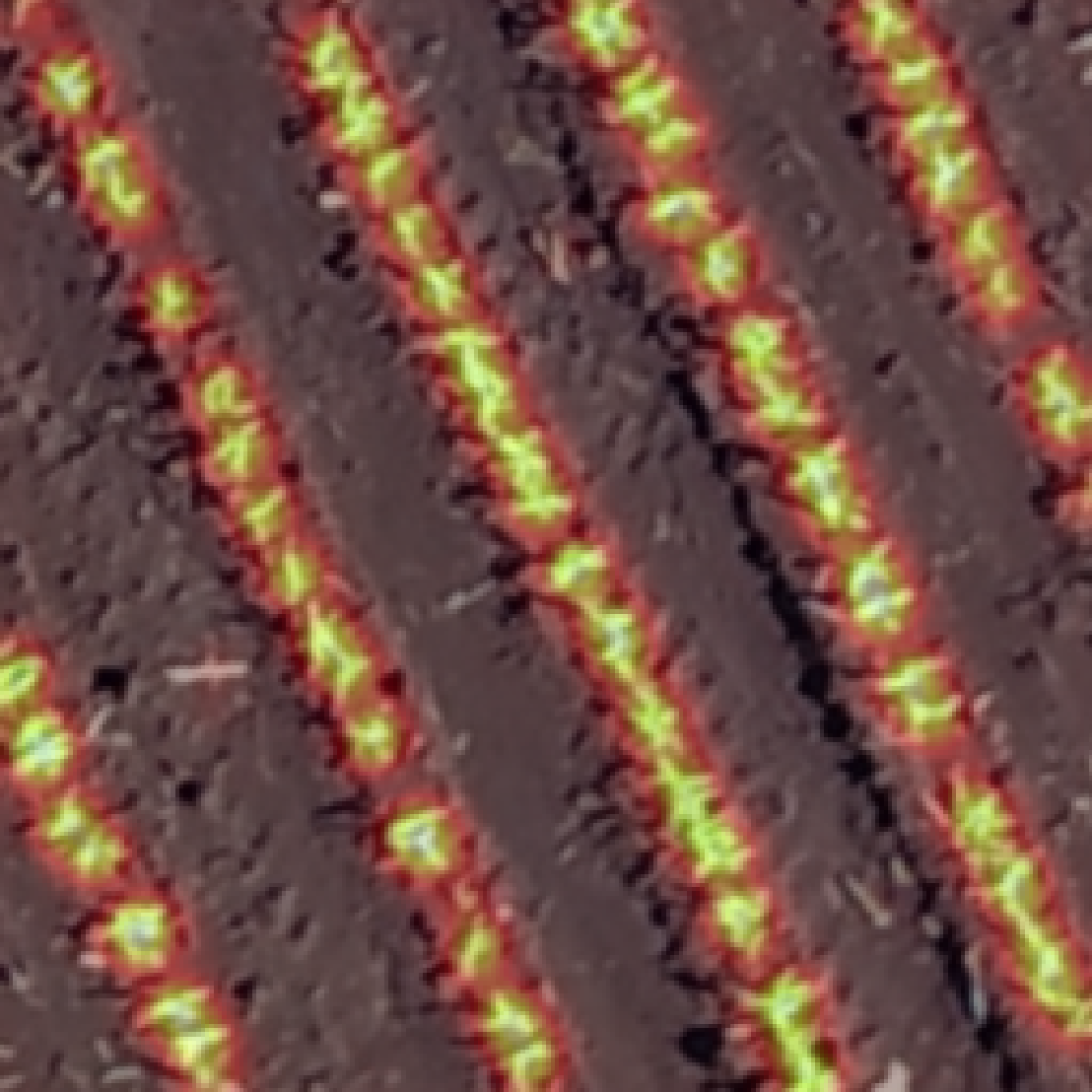}}
	\subfigure{\includegraphics[width=.3\columnwidth]{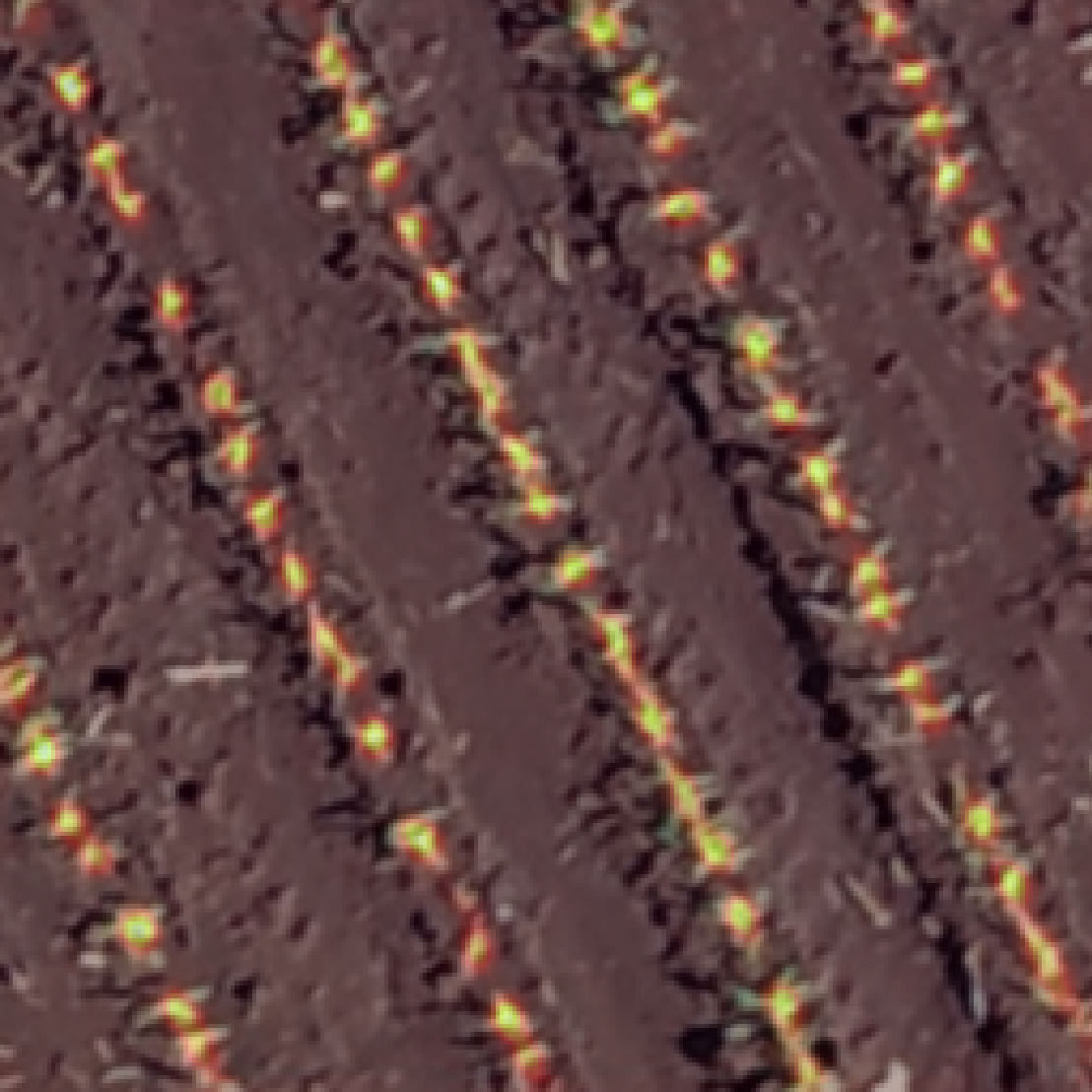}}
	
	\subfigure{\includegraphics[width=.3\columnwidth]{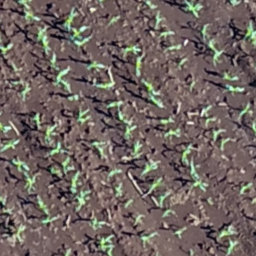}}
	\subfigure{\includegraphics[width=.3\columnwidth]{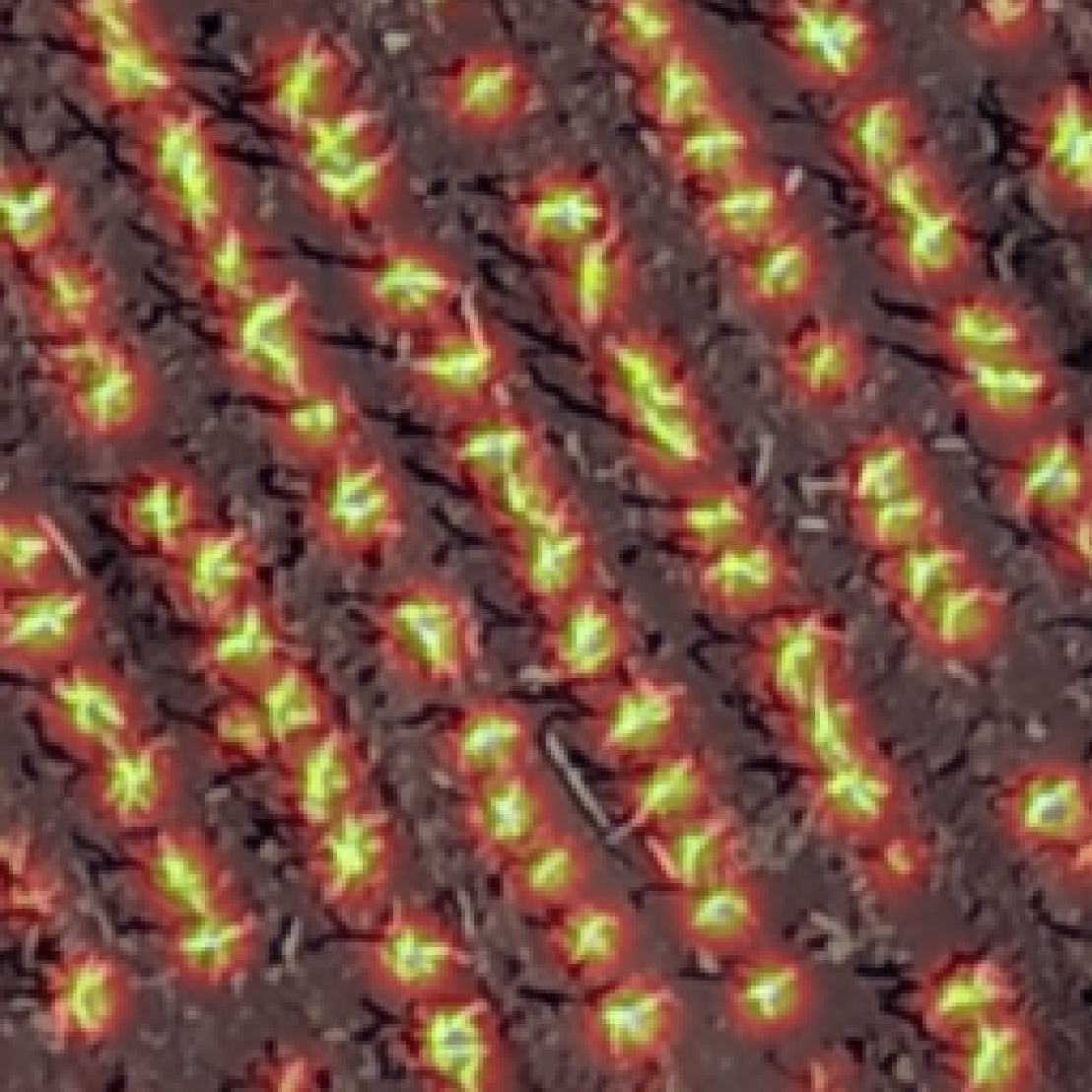}}
	\subfigure{\includegraphics[width=.3\columnwidth]{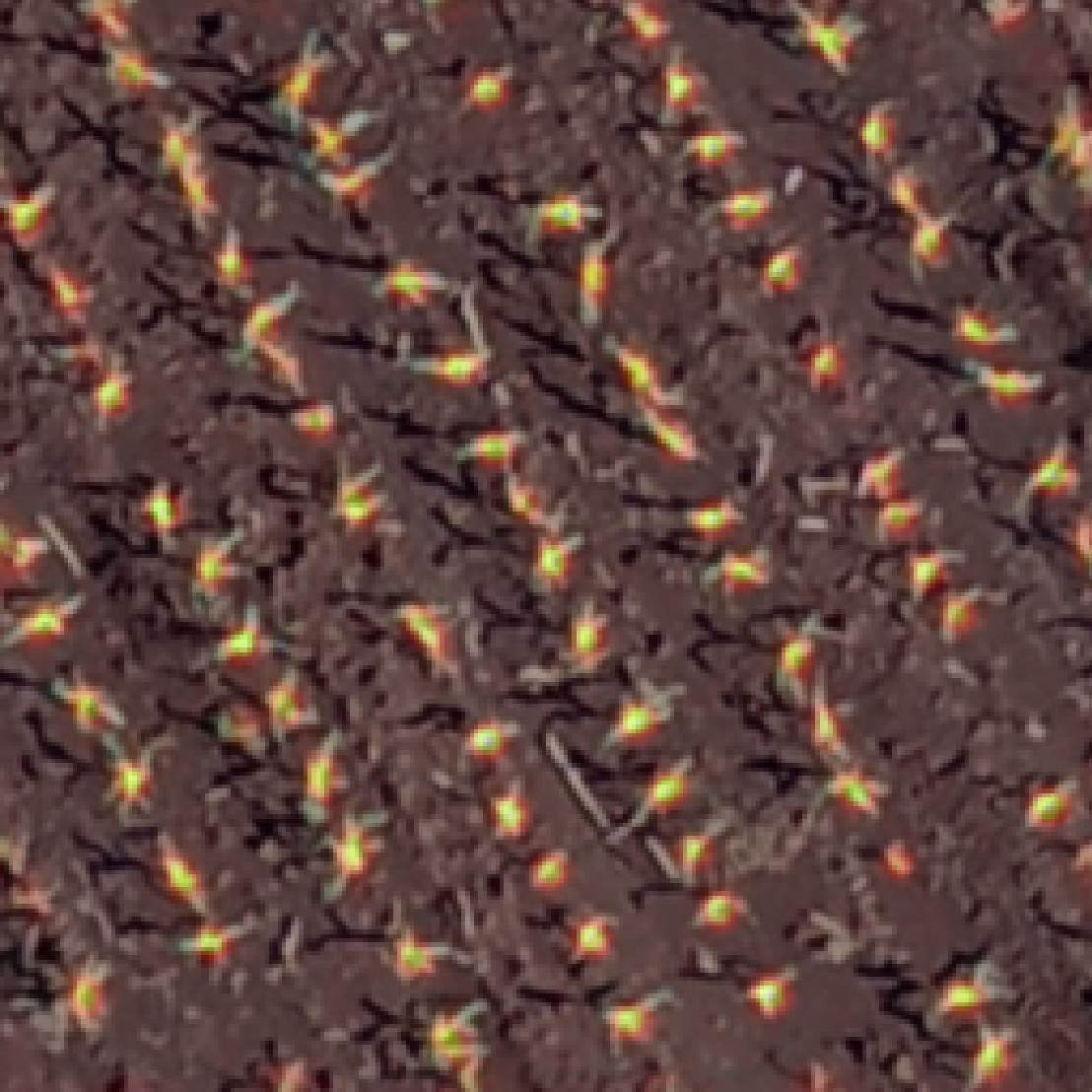}}
	
%
	
	\setcounter{subfigure}{0}
	\subfigure[RGB Image]{\includegraphics[width=.3\columnwidth]{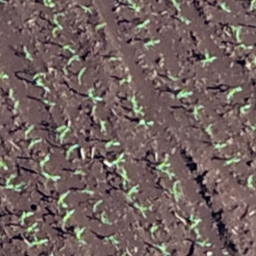}}
	\subfigure[First stage]{\includegraphics[width=.3\columnwidth]{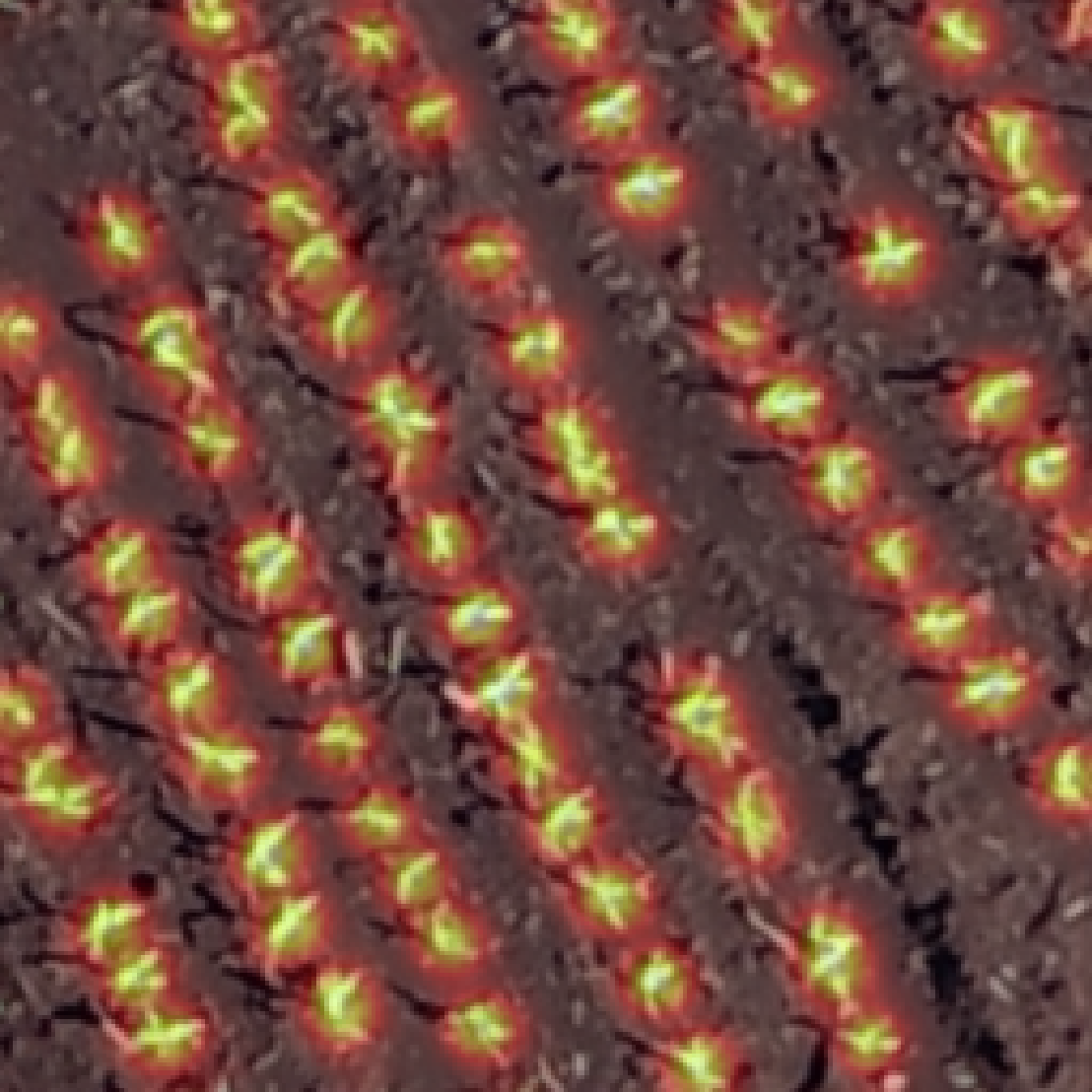}}
	\subfigure[Second stage]{\includegraphics[width=.3\columnwidth]{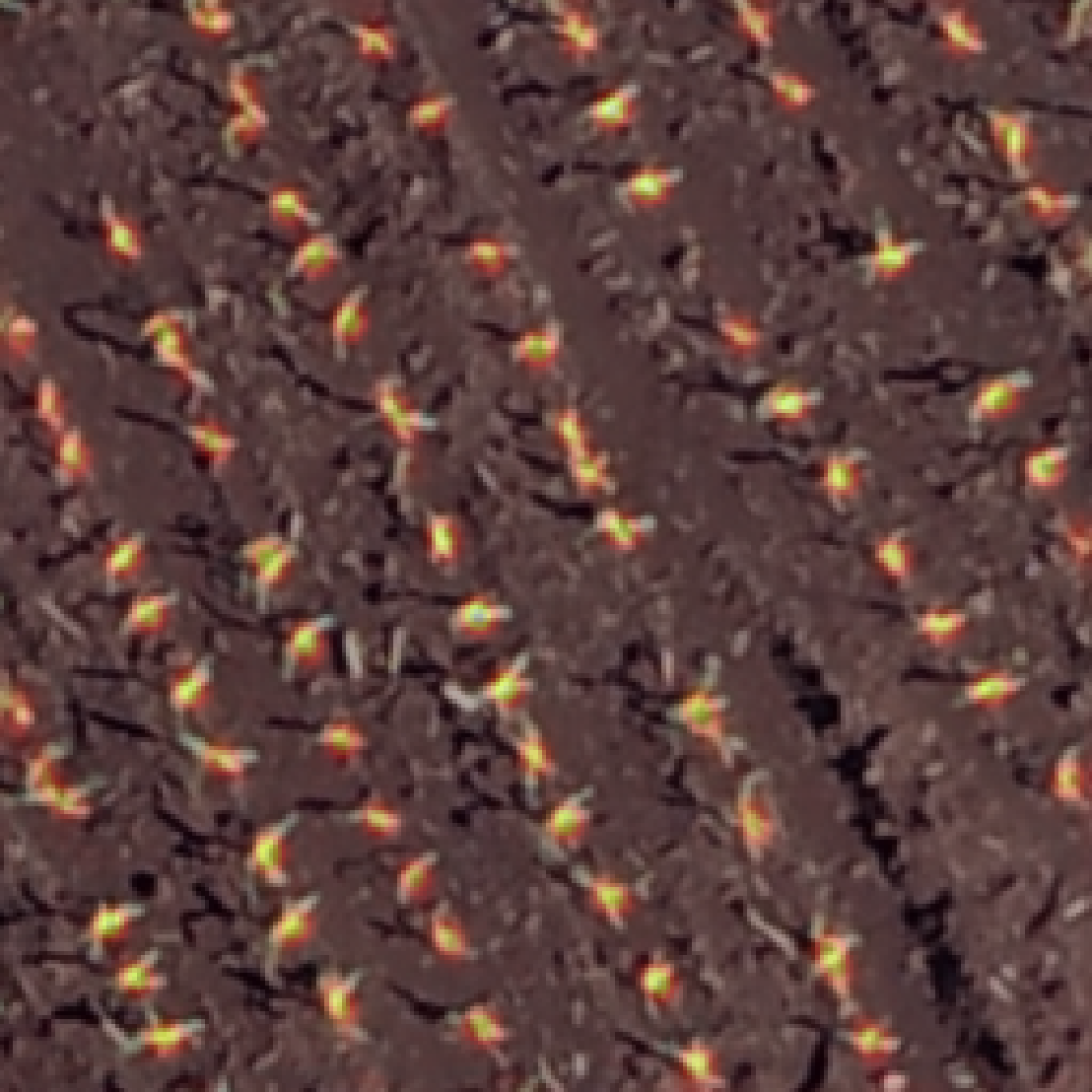}}
	
%
	
	
	\caption{Confidence map of the first and second stages for plant detection.}
	\label{fig:estagios}
\end{figure}

We can see that by increasing the number of stages from 1 to 2, a significant improvement is obtained in the plant detection(e.g., F1 from 0.843 to 0.915). On the other hand, the results stabilize with the number of stages above 2, showing that two stages are sufficient for this step. This is because when using two or more stages, the proposed method is able to refine the detection of the first stage. Figure \ref{fig:estagios} shows the confidence map of the first and second KEM stages for three images of the test set. It is possible to notice that the second stage provides a refinement in the plant detection, which reflects an improvement since two nearby plants can be detected separately.

\begin{table}[ht]
	\centering
	\begin{tabular}{|c|c|c|c|c|}
		\hline
		\textbf{Stages} & \textbf{MAE} & \textbf{Precision(\%)} & \textbf{Recall(\%)} & \textbf{F1(\%)} \\
		\hline
		1 & 10.221 & 78.9 & 91.0 & 84.3 \\
		\hline
		\textbf{2} & \textbf{3.531} & \textbf{92.7} & \textbf{90.5} & \textbf{91.5} \\
		\hline
		4 & 3.478 & 91.0 & 91.4 & 91.0 \\
		\hline
		6 & 3.495 & 91.4 & 90.9 & 91.0 \\
		\hline
		8 & 3.885 & 89.5 & 92.0 & 90.6 \\
		\hline
	\end{tabular}
	\caption{Evaluation of the number of stages in the plant detection.}
	\label{tab:estagiosmilho}
\end{table}

\begin{figure}[ht]
	\centering
	\subfigure{\includegraphics[width=.49\columnwidth]{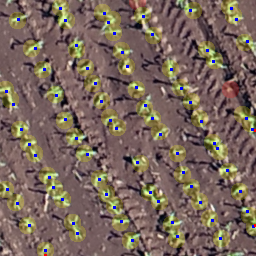}}
	\subfigure{\includegraphics[width=.49\columnwidth]{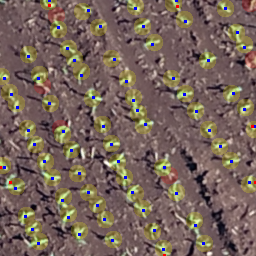}}
	\caption{Examples of plant detection. Blue dots mean correctly predicted plants, red dots are false positives and red circles are false negatives.}
	\label{fig:detecmilho}
\end{figure}

Examples of plant detection are shown in Figure \ref{fig:detecmilho}. In these figures, a correctly predicted plant (True Positive) is illustrated as a blue dot. The red dots represent false positives, that is, detections that are not plants. Plants that were labeled but were not detected by the method are shown by red circles (the radius of the circle corresponds to the metric threshold). The method is able to detect the vast majority of plants, although it fails to detect some plants very close to each other.

Despite this step, obtaining good results (Precision, Recall, and F1 score of 92.7\%, 90.5\%, and 91.5\%), the detection of all plants in the image is not necessary for the correct detection of the plantation lines. However, the more robust the plant detection is, the greater the chance that the line will be detected correctly.

\subsubsection{Edge Classification}

The edge classification module extracts information from the backbone and KEM using $L$ equidistant points along the edge. Then the edges are classified and the plantation lines can be detected. The quantitative assessment of the number of sampled points is shown in Table \ref{tab:l}. When few points are sampled (e.g., $L=4$), the features extracted are insufficient to describe the information, especially when two plants are spatially distant in the image.
On the other hand, $L \geq 8$ presents satisfactory results for images with a resolution of $256 \times 256$ pixels. The best results were obtained with $L=16$, reaching F1 score of 95.1\%.

\begin{table}[ht]
	\centering
	\begin{tabular}{|c|c|c|c|}
		\hline
		\textbf{Number of points} & \textbf{Precision(\%)} & \textbf{Recall(\%)} & \textbf{F1(\%)} \\
		\hline
		4 & 52.4 ($\pm$37.2) & 11.2 ($\pm$09.7) & 16.8 ($\pm$13.7) \\
		\hline
		8 & 98.5 ($\pm$01.8) & 91.0 ($\pm$05.3) & 94.5 ($\pm$03.6) \\
		\hline
		12 & 98.5 ($\pm$01.8) & 91.5 ($\pm$04.8) & 94.7 ($\pm$03.3) \\
		\hline
		16 & 98.7 ($\pm$01.6)	& 91.9 ($\pm$04.3) & 95.1 ($\pm$02.9)  \\
		\hline
		20 & 98.6 ($\pm$01.8) & 91.9 ($\pm$04.3) & 95.0 ($\pm$02.9) \\
		\hline
	\end{tabular}
	\caption{Evaluation of the number of sampled points $L$ in the detection plantation lines.}
	\label{tab:l}
\end{table}

\subsubsection{Combined Information in the Plantation Line Detection}

The edge classification module considers three features to classify an edge as a plantation line: visual, line, and displacement vector features. To assess the influence of each feature, Table \ref{tab:infos} presents the results considering different combinations of features for the edge classification.

When using only the visual features from the backbone, the results are satisfactory with an F1 of 90.7\%. When visual features are combined with line or displacement vector features, F1 is increased to 92.3\% and 94.9\%, respectively. This shows that the features estimated by the KEM are important and assist in the detection of plantation lines. Furthermore, by combining the features as proposed in this work, the best result is obtained.

\begin{table}[ht]
	\centering
	\begin{tabular}{|c|c|c|c|}
		\hline
		\textbf{Features} & \textbf{Precision(\%)} & \textbf{Recall(\%)} & \textbf{F1(\%)} \\
		\hline
		Visual Features & 94.7 ($\pm$06.0) & 87.5 ($\pm$09.5) & 90.7 ($\pm$07.8) \\ \hline
		Visual + Vector Features & 96.3 ($\pm$04.4) & 89.0 ($\pm$07.9) & 92.3 ($\pm$06.2) \\ \hline
		Visual + Line Features & 98.4 ($\pm$01.9) & 91.9 ($\pm$04.3) & 94.9 ($\pm$02.9) \\ \hline
		All Features & 98.7 ($\pm$01.6)	& 91.9 ($\pm$04.3) & 95.1 ($\pm$02.9)  \\
		\hline
	\end{tabular}
	\caption{Results obtained for different combinations of the features used in the edge classification module.}
	\label{tab:infos}
\end{table}

\begin{figure}[ht]
	\centering
	\subfigure{\includegraphics[width=.19\columnwidth]{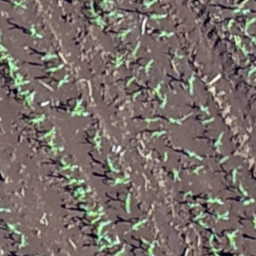}}
	\subfigure{\includegraphics[width=.19\columnwidth]{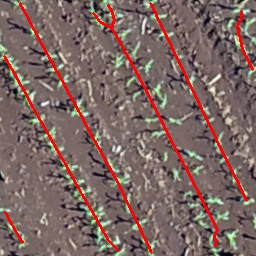}}
	\subfigure{\includegraphics[width=.19\columnwidth]{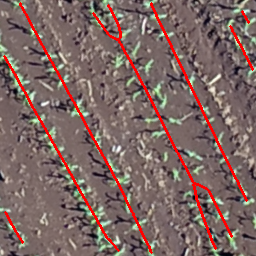}}
	\subfigure{\includegraphics[width=.19\columnwidth]{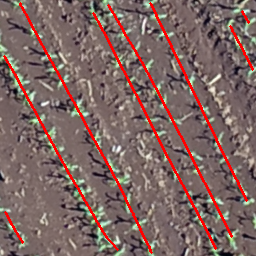}}
	\subfigure{\includegraphics[width=.19\columnwidth]{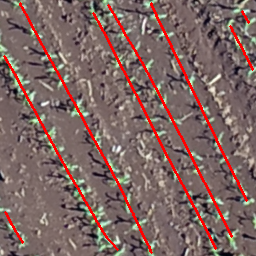}}
	
	\subfigure{\includegraphics[width=.19\columnwidth]{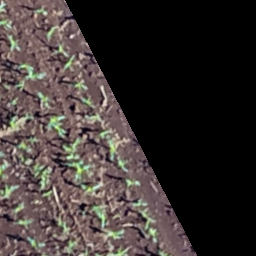}}
	\subfigure{\includegraphics[width=.19\columnwidth]{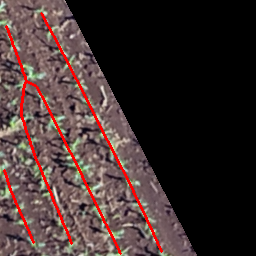}}
	\subfigure{\includegraphics[width=.19\columnwidth]{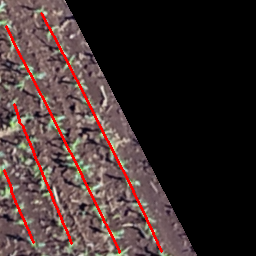}}
	\subfigure{\includegraphics[width=.19\columnwidth]{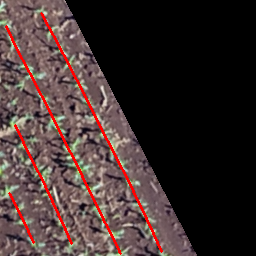}}
	\subfigure{\includegraphics[width=.19\columnwidth]{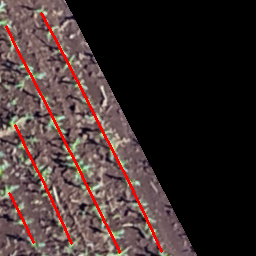}}
	
	\setcounter{subfigure}{0}
	\subfigure[]{\label{fig:infosa}\includegraphics[width=.19\columnwidth]{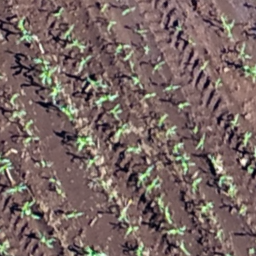}}
	\subfigure[]{\label{fig:infosb}\includegraphics[width=.19\columnwidth]{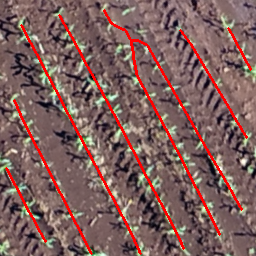}}
	\subfigure[]{\label{fig:infosc}\includegraphics[width=.19\columnwidth]{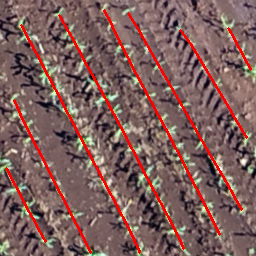}}
	\subfigure[]{\label{fig:infosd}\includegraphics[width=.19\columnwidth]{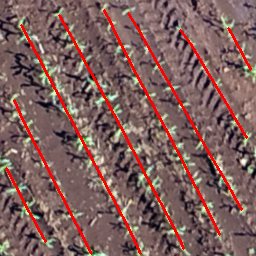}}
	\subfigure[]{\label{fig:infose}\includegraphics[width=.19\columnwidth]{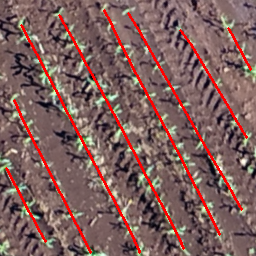}}
	
	\caption{Examples of plantation line detection considering different combinations of features in the edge classification module. (a) RGB image, (b) Visual feature, (c) Visual + Vector displacement features, (d) Visual + line features, (e) All features.}
	\label{fig:infos}
\end{figure}

Examples of plantation line detection are presented in Figure \ref{fig:infos}. Figure \ref{fig:infosa} presents the RGB image of three examples, while Figures \ref{fig:infosb}, \ref{fig:infosc}, \ref{fig:infosd}, and \ref{fig:infose} present the detection using visual features, visual + vector displacement features, visual + line features, and all features, respectively. The main challenges occur when two plantation lines are very close. The first example shows that the visual features and the visual + displacement vector features joined two lines in a single one while the other combinations of features were able to detect them independently. The second and third examples show that the visual features ended up joining two lines at the end, which did not happen with the other combinations. This is because the visual features do not extract structural and shape information, making two plants close in any direction a plausible connection.

\subsection{Comparison with State-of-the-Art Methods}

The proposed method was compared with two recent state-of-the-art methods in Table \ref{tab:sota}. Deep Hough Transform \citep{lin2020deep} integrated the classical Hough transform into deeply learned representations, obtaining promising results in line detection using public datasets. PPGNet \citep{Zhang_2019_CVPR} is similar to the proposed method since it models the problem as a graph. However, PPGNet uses only visual information to classify an edge, in addition to classifying the entire adjacency matrix, which results in a high computational cost. To address this issue, PPGNet performs block prediction to classify the whole adjacency matrix. It is important to emphasize that none of these previous methods has been applied to detect plantation lines.

\begin{figure}
	\centering
	\subfigure{\includegraphics[width=.24\columnwidth]{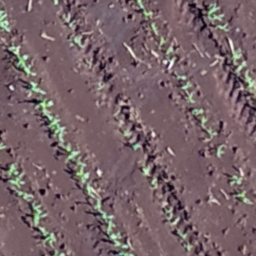}}
	\subfigure{\includegraphics[width=.24\columnwidth]{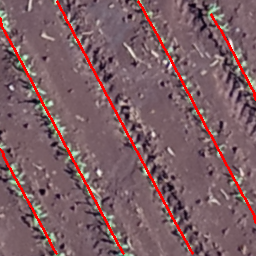}}
	\subfigure{\includegraphics[width=.24\columnwidth]{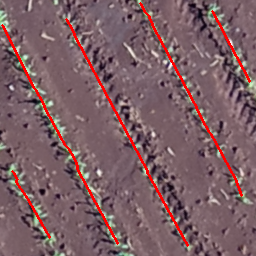}}
	\subfigure{\includegraphics[width=.24\columnwidth]{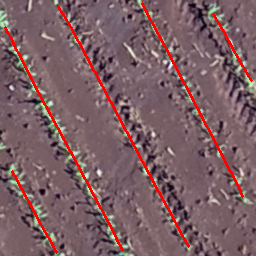}}
	
	\subfigure{\includegraphics[width=.24\columnwidth]{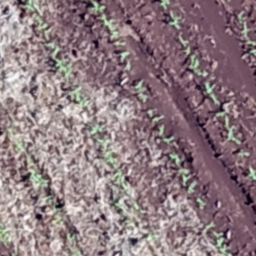}}
	\subfigure{\includegraphics[width=.24\columnwidth]{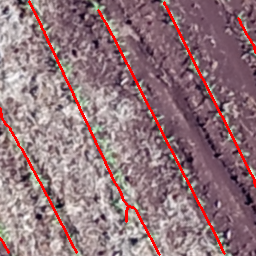}}
	\subfigure{\includegraphics[width=.24\columnwidth]{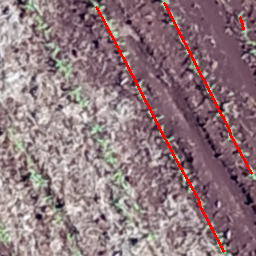}}
	\subfigure{\includegraphics[width=.24\columnwidth]{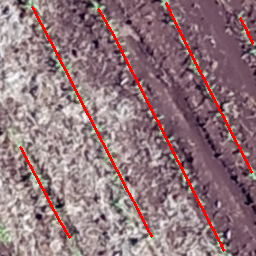}}
	
	\subfigure{\includegraphics[width=.24\columnwidth]{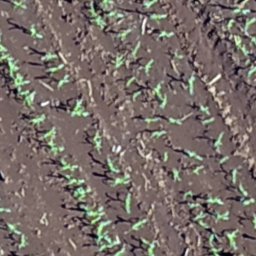}}
	\subfigure{\includegraphics[width=.24\columnwidth]{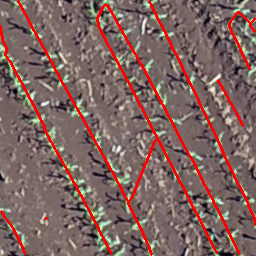}}
	\subfigure{\includegraphics[width=.24\columnwidth]{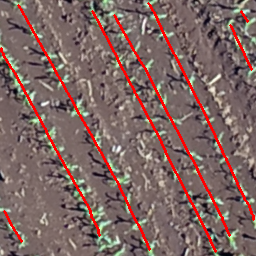}}
	\subfigure{\includegraphics[width=.24\columnwidth]{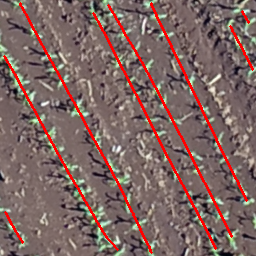}}
	
	\setcounter{subfigure}{0}
	\subfigure[RGB Image]{\includegraphics[width=.24\columnwidth]{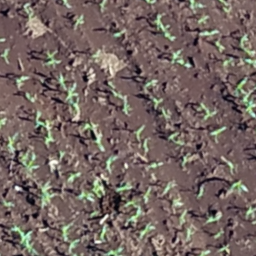}}
	\subfigure[PPGNet]{\includegraphics[width=.24\columnwidth]{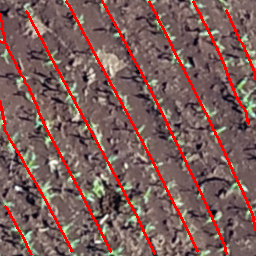}}
	\subfigure[Deep Hough Transform]{\includegraphics[width=.24\columnwidth]{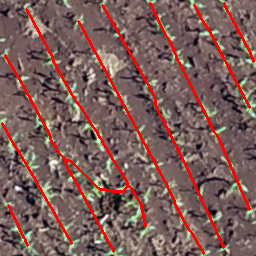}}
	\subfigure[Proposed Method]{\includegraphics[width=.24\columnwidth]{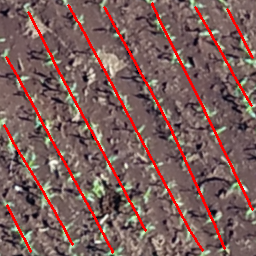}}
	
	\caption{Examples of plantation line detection obtained by the compared methods.}
	\label{fig:sota}
\end{figure}

Experimental results indicate that the proposed method significantly improves F1 score over the traditional approaches, from 91.0\% to 95.1\%. The same occurs for precision and recall, whose best values were obtained by the proposed method. This shows that the use of additional information (e.g., displacement vectors and line pixel probability) can lead to an improvement in the description of the problem. All methods show good results when the plantation lines are well defined as in the first example of Figure \ref{fig:sota}. On the other hand, Deep Hough Transform has difficulty in detecting lines in regions whose plants are not completely visible (see the second example in Figure \ref{fig:sota}). In addition, some examples have shown that state-of-the-art methods connect different plantation lines (the last two examples in the figure). Hence, the method described here has proven to be effective for plantation line detection.

\begin{table}[ht]
	\centering
	\begin{tabular}{|c|c|c|c|}
		\hline
		\textbf{Stages} & \textbf{Precision(\%)} & \textbf{Recall(\%)} & \textbf{F1(\%)} \\
		\hline
		Deep Hough Transform \citep{lin2020deep} & 94.7 ($\pm$06.4) & 87.5 ($\pm$09.9) & 90.1 ($\pm$08.7) \\ \hline
		PPGNet \citep{Zhang_2019_CVPR} & 95.0 ($\pm$03.5) & 87.6 ($\pm$05.5) & 91.0 ($\pm$03.5) \\ \hline
		Proposed Method & 98.7 ($\pm$01.6) & 91.9 ($\pm$04.3) & 95.1 ($\pm$02.9)  \\
		\hline
	\end{tabular}
	\caption{Comparison of the proposed method with two recent state-of-the-art methods.}
	\label{tab:sota}
\end{table}

\section{Discussion}

In this study, we investigated the performance of a deep neural network in combination with the graph theory to extract plantation-lines in RGB images to attend agricultural farmlands. For this, we demonstrated the application of our approach in a corn field dataset composed of corn plants at different growth stages and with different plantation patterns (i.e., directions, curves, space inbetween, etc.). The results from our experiment demonstrated that the proposed approach is feasible to detect both plant and lines positions with high accuracy. Moreover, the comparison of our method against \citep{lin2020deep} and \citep{Zhang_2019_CVPR} deep neural networks indicated that our method is capable of returning accurate results, better than those of the state-of-the-art, and, when compared against its baseline (Visual Features), an improvement from 0.907 to 0.951 occurred. As such, we intend to discuss here this improvement and the importance of graphs theory in conjunction with DNN model.

In our approach, we initially identified the plants' position in the image through a confidence map, being this information useful for estimating the plantation lines. Then, the probability that a pixel belongs to a crop line is estimated and, finally, is estimated the displacement of vectors linking one plant to another on the same plantation line. After these estimates, the problem of detecting plantation lines is modeled using a graph, in which each plant identified in the confidence map is assumed as a vertex in the graph, and these vertices are connected forming a complete graph. Each edge between two vertices, then, is used in the edge classification module to classify whether the edges are a planting line. During the aforementioned process, we verified that at the second stage of the KEM the networks' performance works better and that increasing this number of stages would only result in worse results and higher processing time. After this, the plants, which are viewed as the ''vertices'' by the model, are classified using a given distance between points, where the plantation lines are determined. This information is important since the plantation-line is detected by considering both visual aspects (i.e. spectral and spatial features, texture, pattern, etc.) the line shape itself, and the displacement of the vector features. By considering this displacement of the graph's structure, the network is capable of improving its learning capability concerning the line pattern, especially when differences in the terrain or the direction of the line occurs, since it accounts for the plants' (i.e., vertices) position to one another.

The adoption of graphs theory in deep learning-related approaches is a relatively new concept in remote sensing and has been explored majorly in semantic segmentation tasks \citep{Ouyang2021, Hong2020, Yan2019, 9210789}. These studies mostly investigated graph convolutional networks and attention-based mechanisms, which differs from the proposal present here. Regardless, there is no denying the graph addition has the potential to assist in learning patterns and positions of most of the surfaces' targets. In remote sensing applied to agricultural problems, the integration with graphs can help ascertain a series of object detection tasks, especially those that involve certain patterns and geometry information, as any of other anthropic based environments. As such, this approach offers potential not only for plantation line detection but also for other linear forms like river and its margins, roads and side-roads, sidewalks, utility pole-lines, among others, which were already the theme of previous deep learning approaches related to both segmentation and object detection \citep{Weld2019, Gomes2020, Weng2020, Yang2019}

The detection of plantation lines is not an easy task to be performed by automatic methods, and the usage of graph is necessary to assist it. Some challenges that occurred when considering our baseline, which only considered the visual features and the first two information branches to rely on the plantation lines' position, was the presence of plants outside the plantation lines' range (i.e. highly spaced gaps), as well as isolated plants and weeds, that both offered a hindrance to the plantation-line detection process. Here, when considering the third information branch with the displacement of the vector features, most of these problems were dealt with, resulting in its better performance, both visually and numerically. Regardless, previously conducted approaches that intended to extract plantation lines from aerial RGB imagery were also reportedly successfully, specifically to detect citrus-tress planted in curved rows \citep{Rosa2020}, which form intricate geometric patterns in the image, as well as in an unsupervised manner, in which the plantation line segmentation was a complementary approach to detect weeds outside the line \citep{DianBah2018}.

Future perspectives on graph application in combination with deep convolutional neural networks (or any other type of network) for remote sensing approaches should be encouraged. Deep networks are a powerful method for extracting and learning patterns in imagery. However, they tend to ignore the basic principles of the object pattern in the real world. Graphs, on the other hand, can represent these features and their relationship accordingly. As such, this combination of knowledge provided by both methods is quickly gaining attention in remote sensing and photograpmetric field, where most real-life patterns are represented. In this regard, discovering learning patterns related to automatic agricultural practices, such as extracting plantation-line information, is one of the many types of geometric-related mappings that could be potentially benefited from the addition of graphs into the DNN model. In summary, our approach demonstrated that the network improved its performance when considering this novel information into its learning process by achieving better accuracies than its previous structure and other state-of-the-art methods, as aforementioned.

\section{Conclusion}
This paper presents a novel deep learning-based method to extract plantation lines in aerial imagery of agricultural fields. Our approach extracts knowledge from the feature map organized in three extraction and refinement branches for plant positions, plantation lines, and for the displacement vectors between the plants. A graph modeling is applied considering each plant as a vertex, and the edges are formed between two plants. As the edge is classified as belonging to a certain plantation line based on visual learning features extracted from the backbone, our approach enhances this since there is also a chance that the plant pixel belongs to a line, which is extracted by the KEM method and is refined with information from the alignment of the displacement vectors with the plant/object. Based on the experiments, our approach can be characterized as an effective strategy for dealing with hard-to-detect lines, especially those with spaced plants. When it was compared against the state-of-the-art deep learning methods, including Deep Hough Transform and PPGNet, our approach demonstrated superior performance with a significant margin, returning precision, recall, and F1 scores of 98.7\%, 91.9\%, and 95.1\%, respectively, for our method. Therefore, it represents an innovative strategy for extracting lines with spaced plantation patterns, and it could be implemented in scenarios where plantation gaps occur, generating lines with few-to-none interruptions.

\section*{Acknowledgments.}
This study was supported by the FUNDECT - State of Mato Grosso do Sul Foundation to Support Education, Science and Technology, CAPES - Brazilian Federal Agency for Support and Evaluation of Graduate Education, and CNPq - National Council for Scientific and Technological Development. The Titan V and XP used for this research was donated by the NVIDIA Corporation.

\bibliographystyle{unsrtnat}


\begin{thebibliography}{42}
\expandafter\ifx\csname natexlab\endcsname\relax\def\natexlab#1{#1}\fi
\providecommand{\url}[1]{\texttt{#1}}
\providecommand{\href}[2]{#2}
\providecommand{\path}[1]{#1}
\providecommand{\DOIprefix}{doi:}
\providecommand{\ArXivprefix}{arXiv:}
\providecommand{\URLprefix}{URL: }
\providecommand{\Pubmedprefix}{pmid:}
\providecommand{\doi}[1]{\href{http://dx.doi.org/#1}{\path{#1}}}
\providecommand{\Pubmed}[1]{\href{pmid:#1}{\path{#1}}}
\providecommand{\bibinfo}[2]{#2}
\ifx\xfnm\relax \def\xfnm[#1]{\unskip,\space#1}\fi
\bibitem[{Babapour et~al.(2017)Babapour, Mokhtarzade and Zoej}]{Badapour2017}
\bibinfo{author}{Babapour, H.}, \bibinfo{author}{Mokhtarzade, M.},
  \bibinfo{author}{Zoej, M.J.V.}, \bibinfo{year}{2017}.
\newblock \bibinfo{title}{A novel post-calibration method for digital cameras
  using image linear features}.
\newblock \bibinfo{journal}{International Journal of Remote Sensing}
  \bibinfo{volume}{38}, \bibinfo{pages}{2698--2716}.
\newblock \DOIprefix\doi{10.1080/01431161.2016.1232875}.
\bibitem[{Badrinarayanan et~al.(2017)Badrinarayanan, Kendall and
  Cipolla}]{Badrinarayanan2017}
\bibinfo{author}{Badrinarayanan, V.}, \bibinfo{author}{Kendall, A.},
  \bibinfo{author}{Cipolla, R.}, \bibinfo{year}{2017}.
\newblock \bibinfo{title}{{SegNet: A Deep Convolutional Encoder-Decoder
  Architecture for Image Segmentation}}.
\newblock \bibinfo{journal}{IEEE Transactions on Pattern Analysis and Machine
  Intelligence} \bibinfo{volume}{39}, \bibinfo{pages}{2481--2495}.
\newblock \DOIprefix\doi{10.1109/TPAMI.2016.2644615},
  \href{http://arxiv.org/abs/1511.00561}{\tt arXiv:1511.00561}.
\bibitem[{{Cai} and {Wei}(2020)}]{9210789}
\bibinfo{author}{{Cai}, W.}, \bibinfo{author}{{Wei}, Z.}, \bibinfo{year}{2020}.
\newblock \bibinfo{title}{Remote sensing image classification based on a
  cross-attention mechanism and graph convolution}.
\newblock \bibinfo{journal}{IEEE Geoscience and Remote Sensing Letters} ,
  \bibinfo{pages}{1--5}\DOIprefix\doi{10.1109/LGRS.2020.3026587}.
\bibitem[{{Cao} et~al.(2017){Cao}, {Simon}, {Wei} and {Sheikh}}]{Cao2017}
\bibinfo{author}{{Cao}, Z.}, \bibinfo{author}{{Simon}, T.},
  \bibinfo{author}{{Wei}, S.}, \bibinfo{author}{{Sheikh}, Y.},
  \bibinfo{year}{2017}.
\newblock \bibinfo{title}{Realtime multi-person 2d pose estimation using part
  affinity fields}, in: \bibinfo{booktitle}{2017 IEEE Conference on Computer
  Vision and Pattern Recognition (CVPR)}, pp. \bibinfo{pages}{1302--1310}.
\newblock \DOIprefix\doi{10.1109/CVPR.2017.143}.
\bibitem[{{Dian Bah} et~al.(2018){Dian Bah}, Hafiane and Canals}]{DianBah2018}
\bibinfo{author}{{Dian Bah}, M.}, \bibinfo{author}{Hafiane, A.},
  \bibinfo{author}{Canals, R.}, \bibinfo{year}{2018}.
\newblock \bibinfo{title}{{Deep learning with unsupervised data labeling for
  weed detection in line crops in UAV images}}.
\newblock \bibinfo{journal}{Remote Sensing} \bibinfo{volume}{10},
  \bibinfo{pages}{1--22}.
\newblock \DOIprefix\doi{10.3390/rs10111690}.
\bibitem[{Gao et~al.(2021)Gao, Shi, Li and Wang}]{Gao2021}
\bibinfo{author}{Gao, Y.}, \bibinfo{author}{Shi, J.}, \bibinfo{author}{Li, J.},
  \bibinfo{author}{Wang, R.}, \bibinfo{year}{2021}.
\newblock \bibinfo{title}{{Remote sensing scene classification based on
  high-order graph convolutional network}}.
\newblock \bibinfo{journal}{European Journal of Remote Sensing}
  \bibinfo{volume}{00}, \bibinfo{pages}{1--15}.
\newblock \URLprefix \url{https://doi.org/10.1080/22797254.2020.1868273},
  \DOIprefix\doi{10.1080/22797254.2020.1868273}.
\bibitem[{Gomes et~al.(2020)Gomes, Silva, Gon{\c{c}}alves, Zamboni, Perez,
  Batista, Ramos, Osco, Matsubara, Li, Junior and Gon{\c{c}}alves}]{Gomes2020}
\bibinfo{author}{Gomes, M.}, \bibinfo{author}{Silva, J.},
  \bibinfo{author}{Gon{\c{c}}alves, D.}, \bibinfo{author}{Zamboni, P.},
  \bibinfo{author}{Perez, J.}, \bibinfo{author}{Batista, E.},
  \bibinfo{author}{Ramos, A.}, \bibinfo{author}{Osco, L.},
  \bibinfo{author}{Matsubara, E.}, \bibinfo{author}{Li, J.},
  \bibinfo{author}{Junior, J.M.}, \bibinfo{author}{Gon{\c{c}}alves, W.},
  \bibinfo{year}{2020}.
\newblock \bibinfo{title}{{Mapping utility poles in aerial orthoimages using
  atss deep learning method}}.
\newblock \bibinfo{journal}{Sensors (Switzerland)} \bibinfo{volume}{20},
  \bibinfo{pages}{1--14}.
\newblock \DOIprefix\doi{10.3390/s20216070}.
\bibitem[{Habib et~al.(2005)Habib, Ghanma, Morgan and Al-Ruzouq}]{Habib2005}
\bibinfo{author}{Habib, A.}, \bibinfo{author}{Ghanma, M.},
  \bibinfo{author}{Morgan, M.}, \bibinfo{author}{Al-Ruzouq, R.},
  \bibinfo{year}{2005}.
\newblock \bibinfo{title}{{Photogrammetric and lidar data registration using
  linear features}}.
\newblock \bibinfo{journal}{Photogrammetric Engineering and Remote Sensing}
  \bibinfo{volume}{71}, \bibinfo{pages}{699--707}.
\newblock \DOIprefix\doi{10.14358/PERS.71.6.699}.
\bibitem[{{He} et~al.(2016){He}, {Zhang}, {Ren} and {Sun}}]{Resnet16}
\bibinfo{author}{{He}, K.}, \bibinfo{author}{{Zhang}, X.},
  \bibinfo{author}{{Ren}, S.}, \bibinfo{author}{{Sun}, J.},
  \bibinfo{year}{2016}.
\newblock \bibinfo{title}{Deep residual learning for image recognition}, in:
  \bibinfo{booktitle}{2016 IEEE Conference on Computer Vision and Pattern
  Recognition (CVPR)}, pp. \bibinfo{pages}{770--778}.
\newblock \DOIprefix\doi{10.1109/CVPR.2016.90}.
\bibitem[{Hong et~al.(2020)Hong, Gao, Yao, Zhang, Plaza and
  Chanussot}]{Hong2020}
\bibinfo{author}{Hong, D.}, \bibinfo{author}{Gao, L.}, \bibinfo{author}{Yao,
  J.}, \bibinfo{author}{Zhang, B.}, \bibinfo{author}{Plaza, A.},
  \bibinfo{author}{Chanussot, J.}, \bibinfo{year}{2020}.
\newblock \bibinfo{title}{{Graph convolutional networks for hyperspectral image
  classification}}.
\newblock \bibinfo{journal}{arXiv} ,
  \bibinfo{pages}{1--13}\DOIprefix\doi{10.1109/tgrs.2020.3015157},
  \href{http://arxiv.org/abs/2008.02457}{\tt arXiv:2008.02457}.
\bibitem[{Kubik(1991)}]{KUBIK1991}
\bibinfo{author}{Kubik, K.}, \bibinfo{year}{1991}.
\newblock \bibinfo{title}{Relative and absolute orientation based on linear
  features}.
\newblock \bibinfo{journal}{ISPRS Journal of Photogrammetry and Remote Sensing}
  \bibinfo{volume}{46}, \bibinfo{pages}{199 -- 204}.
\newblock \URLprefix
  \url{http://www.sciencedirect.com/science/article/pii/092427169190053X},
  \DOIprefix\doi{https://doi.org/10.1016/0924-2716(91)90053-X}.
\bibitem[{Lee and Bethel(2004)}]{LEE2004}
\bibinfo{author}{Lee, C.}, \bibinfo{author}{Bethel, J.S.},
  \bibinfo{year}{2004}.
\newblock \bibinfo{title}{Extraction, modelling, and use of linear features for
  restitution of airborne hyperspectral imagery}.
\newblock \bibinfo{journal}{ISPRS Journal of Photogrammetry and Remote Sensing}
  \bibinfo{volume}{58}, \bibinfo{pages}{289 -- 300}.
\newblock \URLprefix
  \url{http://www.sciencedirect.com/science/article/pii/S0924271603000613},
  \DOIprefix\doi{https://doi.org/10.1016/j.isprsjprs.2003.10.003}.
\bibitem[{{Li} and {Shi}(2014)}]{6709786}
\bibinfo{author}{{Li}, C.}, \bibinfo{author}{{Shi}, W.}, \bibinfo{year}{2014}.
\newblock \bibinfo{title}{The generalized-line-based iterative transformation
  model for imagery registration and rectification}.
\newblock \bibinfo{journal}{IEEE Geoscience and Remote Sensing Letters}
  \bibinfo{volume}{11}, \bibinfo{pages}{1394--1398}.
\newblock \DOIprefix\doi{10.1109/LGRS.2013.2293844}.
\bibitem[{Lin et~al.(2020)Lin, Pintea and van Gemert}]{lin2020deep}
\bibinfo{author}{Lin, Y.}, \bibinfo{author}{Pintea, S.L.}, \bibinfo{author}{van
  Gemert, J.C.}, \bibinfo{year}{2020}.
\newblock \bibinfo{title}{Deep hough-transform line priors}.
\newblock \href{http://arxiv.org/abs/2007.09493}{\tt arXiv:2007.09493}.
\bibitem[{Long et~al.(2015a)Long, Shelhamer and Darrell}]{Long_2015_CVPR}
\bibinfo{author}{Long, J.}, \bibinfo{author}{Shelhamer, E.},
  \bibinfo{author}{Darrell, T.}, \bibinfo{year}{2015}a.
\newblock \bibinfo{title}{Fully convolutional networks for semantic
  segmentation}.
\newblock \href{http://arxiv.org/abs/1411.4038}{\tt arXiv:1411.4038}.
\bibitem[{Long et~al.(2015b)Long, Jiao, He, Zhang, Cheng and Wang}]{LONG2015}
\bibinfo{author}{Long, T.}, \bibinfo{author}{Jiao, W.}, \bibinfo{author}{He,
  G.}, \bibinfo{author}{Zhang, Z.}, \bibinfo{author}{Cheng, B.},
  \bibinfo{author}{Wang, W.}, \bibinfo{year}{2015}b.
\newblock \bibinfo{title}{A generic framework for image rectification using
  multiple types of feature}.
\newblock \bibinfo{journal}{ISPRS Journal of Photogrammetry and Remote Sensing}
  \bibinfo{volume}{102}, \bibinfo{pages}{161 -- 171}.
\newblock \URLprefix
  \url{http://www.sciencedirect.com/science/article/pii/S0924271615000337},
  \DOIprefix\doi{https://doi.org/10.1016/j.isprsjprs.2015.01.015}.
\bibitem[{Ma et~al.(2019)Ma, Gao, Sun, Zhou and Hussain}]{Ma2019}
\bibinfo{author}{Ma, F.}, \bibinfo{author}{Gao, F.}, \bibinfo{author}{Sun, J.},
  \bibinfo{author}{Zhou, H.}, \bibinfo{author}{Hussain, A.},
  \bibinfo{year}{2019}.
\newblock \bibinfo{title}{{Attention graph convolution network for image
  segmentation in big SAR imagery data}}.
\newblock \bibinfo{journal}{Remote Sensing} \bibinfo{volume}{11},
  \bibinfo{pages}{1--21}.
\newblock \DOIprefix\doi{10.3390/rs11212586}.
\bibitem[{{Marcato Junior} and Tommaselli(2013)}]{MARCATOJUNIOR2013}
\bibinfo{author}{{Marcato Junior}, J.}, \bibinfo{author}{Tommaselli, A.},
  \bibinfo{year}{2013}.
\newblock \bibinfo{title}{Exterior orientation of cbers-2b imagery using
  multi-feature control and orbital data}.
\newblock \bibinfo{journal}{ISPRS Journal of Photogrammetry and Remote Sensing}
  \bibinfo{volume}{79}, \bibinfo{pages}{219 -- 225}.
\newblock \URLprefix
  \url{http://www.sciencedirect.com/science/article/pii/S0924271613000610},
  \DOIprefix\doi{https://doi.org/10.1016/j.isprsjprs.2013.02.018}.
\bibitem[{Noh et~al.(2015)Noh, Hong and Han}]{Noh2015}
\bibinfo{author}{Noh, H.}, \bibinfo{author}{Hong, S.}, \bibinfo{author}{Han,
  B.}, \bibinfo{year}{2015}.
\newblock \bibinfo{title}{Learning deconvolution network for semantic
  segmentation}.
\newblock \href{http://arxiv.org/abs/1505.043661}{\tt arXiv:1505.043661}.
\bibitem[{Osco et~al.(2021a)Osco, dos Santos~de Arruda, Gonçalves, Dias,
  Batistoti, de~Souza, Gomes, Ramos, de~Castro~Jorge, Liesenberg, Li, Ma,
  Junior and Gonçalves}]{osco2021cnn}
\bibinfo{author}{Osco, L.P.}, \bibinfo{author}{dos Santos~de Arruda, M.},
  \bibinfo{author}{Gonçalves, D.N.}, \bibinfo{author}{Dias, A.},
  \bibinfo{author}{Batistoti, J.}, \bibinfo{author}{de~Souza, M.},
  \bibinfo{author}{Gomes, F.D.G.}, \bibinfo{author}{Ramos, A.P.M.},
  \bibinfo{author}{de~Castro~Jorge, L.A.}, \bibinfo{author}{Liesenberg, V.},
  \bibinfo{author}{Li, J.}, \bibinfo{author}{Ma, L.}, \bibinfo{author}{Junior,
  J.M.}, \bibinfo{author}{Gonçalves, W.N.}, \bibinfo{year}{2021}a.
\newblock \bibinfo{title}{A cnn approach to simultaneously count plants and
  detect plantation-rows from uav imagery}.
\newblock \href{http://arxiv.org/abs/2012.15827}{\tt arXiv:2012.15827}.
\bibitem[{Osco et~al.(2020)Osco, de~Arruda, {Marcato Junior}, da~Silva, Ramos,
  Moryia, Imai, Pereira, Creste, Matsubara, Li and Gon{\c{c}}alves}]{Osco2020}
\bibinfo{author}{Osco, L.P.}, \bibinfo{author}{de~Arruda, M.d.S.},
  \bibinfo{author}{{Marcato Junior}, J.}, \bibinfo{author}{da~Silva, N.B.},
  \bibinfo{author}{Ramos, A.P.M.}, \bibinfo{author}{Moryia, {\'{E}}.A.S.},
  \bibinfo{author}{Imai, N.N.}, \bibinfo{author}{Pereira, D.R.},
  \bibinfo{author}{Creste, J.E.}, \bibinfo{author}{Matsubara, E.T.},
  \bibinfo{author}{Li, J.}, \bibinfo{author}{Gon{\c{c}}alves, W.N.},
  \bibinfo{year}{2020}.
\newblock \bibinfo{title}{{A convolutional neural network approach for counting
  and geolocating citrus-trees in UAV multispectral imagery}}.
\newblock \bibinfo{journal}{ISPRS Journal of Photogrammetry and Remote Sensing}
  \bibinfo{volume}{160}, \bibinfo{pages}{97--106}.
\newblock \URLprefix \url{https://doi.org/10.1016/j.isprsjprs.2019.12.010},
  \DOIprefix\doi{10.1016/j.isprsjprs.2019.12.010}.
\bibitem[{Osco et~al.(2021b)Osco, Junior, Ramos, de~Castro~Jorge, Fatholahi,
  de~Andrade~Silva, Matsubara, Pistori, Gonçalves and Li}]{osco2021DL}
\bibinfo{author}{Osco, L.P.}, \bibinfo{author}{Junior, J.M.},
  \bibinfo{author}{Ramos, A.P.M.}, \bibinfo{author}{de~Castro~Jorge, L.A.},
  \bibinfo{author}{Fatholahi, S.N.}, \bibinfo{author}{de~Andrade~Silva, J.},
  \bibinfo{author}{Matsubara, E.T.}, \bibinfo{author}{Pistori, H.},
  \bibinfo{author}{Gonçalves, W.N.}, \bibinfo{author}{Li, J.},
  \bibinfo{year}{2021}b.
\newblock \bibinfo{title}{A review on deep learning in uav remote sensing}.
\newblock \href{http://arxiv.org/abs/2101.10861}{\tt arXiv:2101.10861}.
\bibitem[{Ouyang and Li(2021)}]{Ouyang2021}
\bibinfo{author}{Ouyang, S.}, \bibinfo{author}{Li, Y.}, \bibinfo{year}{2021}.
\newblock \bibinfo{title}{{Combining deep semantic segmentation network and
  graph convolutional neural network for semantic segmentation of remote
  sensing imagery}}.
\newblock \bibinfo{journal}{Remote Sensing} \bibinfo{volume}{13},
  \bibinfo{pages}{1--22}.
\newblock \DOIprefix\doi{10.3390/rs13010119}.
\bibitem[{{Ravi} et~al.(2018){Ravi}, {Lin}, {Elbahnasawy}, {Shamseldin} and
  {Habib}}]{8340844}
\bibinfo{author}{{Ravi}, R.}, \bibinfo{author}{{Lin}, Y.},
  \bibinfo{author}{{Elbahnasawy}, M.}, \bibinfo{author}{{Shamseldin}, T.},
  \bibinfo{author}{{Habib}, A.}, \bibinfo{year}{2018}.
\newblock \bibinfo{title}{Simultaneous system calibration of a multi-lidar
  multicamera mobile mapping platform}.
\newblock \bibinfo{journal}{IEEE Journal of Selected Topics in Applied Earth
  Observations and Remote Sensing} \bibinfo{volume}{11},
  \bibinfo{pages}{1694--1714}.
\newblock \DOIprefix\doi{10.1109/JSTARS.2018.2812796}.
\bibitem[{Ronneberger et~al.(2015)Ronneberger, Fischer and
  Brox}]{Ronneberger2015}
\bibinfo{author}{Ronneberger, O.}, \bibinfo{author}{Fischer, P.},
  \bibinfo{author}{Brox, T.}, \bibinfo{year}{2015}.
\newblock \bibinfo{title}{{U-net: Convolutional networks for biomedical image
  segmentation}}.
\newblock \bibinfo{journal}{Lecture Notes in Computer Science (including
  subseries Lecture Notes in Artificial Intelligence and Lecture Notes in
  Bioinformatics)} \bibinfo{volume}{9351}, \bibinfo{pages}{234--241}.
\newblock \DOIprefix\doi{10.1007/978-3-319-24574-4_28},
  \href{http://arxiv.org/abs/1505.04597}{\tt arXiv:1505.04597}.
\bibitem[{{Rosa} et~al.(2020){Rosa}, {Oliveira}, {Zortea}, {Gemignani} and
  {Feitosa}}]{Rosa2020}
\bibinfo{author}{{Rosa}, L.E.C.L.}, \bibinfo{author}{{Oliveira}, D.A.B.},
  \bibinfo{author}{{Zortea}, M.}, \bibinfo{author}{{Gemignani}, B.H.},
  \bibinfo{author}{{Feitosa}, R.Q.}, \bibinfo{year}{2020}.
\newblock \bibinfo{title}{Learning geometric features for improving the
  automatic detection of citrus plantation rows in uav images}.
\newblock \bibinfo{journal}{IEEE Geoscience and Remote Sensing Letters} ,
  \bibinfo{pages}{1--5}\DOIprefix\doi{10.1109/LGRS.2020.3024641}.
\bibitem[{Schenk(2004)}]{SCHENK2004}
\bibinfo{author}{Schenk, T.}, \bibinfo{year}{2004}.
\newblock \bibinfo{title}{From point-based to feature-based aerial
  triangulation}.
\newblock \bibinfo{journal}{ISPRS Journal of Photogrammetry and Remote Sensing}
  \bibinfo{volume}{58}, \bibinfo{pages}{315 -- 329}.
\newblock \URLprefix
  \url{http://www.sciencedirect.com/science/article/pii/S092427160400005X},
  \DOIprefix\doi{https://doi.org/10.1016/j.isprsjprs.2004.02.003}.
\bibitem[{Simonyan and Zisserman(2015)}]{VGG15}
\bibinfo{author}{Simonyan, K.}, \bibinfo{author}{Zisserman, A.},
  \bibinfo{year}{2015}.
\newblock \bibinfo{title}{Very deep convolutional networks for large-scale
  image recognition}, in: \bibinfo{booktitle}{International Conference on
  Learning Representations}, p.~\bibinfo{pages}{14}.
\bibitem[{Sun et~al.(2019)Sun, Robson, Scott, Boehm and Wang}]{SUN2019}
\bibinfo{author}{Sun, Y.}, \bibinfo{author}{Robson, S.},
  \bibinfo{author}{Scott, D.}, \bibinfo{author}{Boehm, J.},
  \bibinfo{author}{Wang, Q.}, \bibinfo{year}{2019}.
\newblock \bibinfo{title}{Automatic sensor orientation using horizontal and
  vertical line feature constraints}.
\newblock \bibinfo{journal}{ISPRS Journal of Photogrammetry and Remote Sensing}
  \bibinfo{volume}{150}, \bibinfo{pages}{172 -- 184}.
\newblock \URLprefix
  \url{http://www.sciencedirect.com/science/article/pii/S0924271619300449},
  \DOIprefix\doi{https://doi.org/10.1016/j.isprsjprs.2019.02.011}.
\bibitem[{Tommaselli and Junior(2012)}]{Tommaselli2012}
\bibinfo{author}{Tommaselli, A.M.}, \bibinfo{author}{Junior, J.M.},
  \bibinfo{year}{2012}.
\newblock \bibinfo{title}{Bundle block adjustment of cbers-2b hrc imagery
  combining control points and lines}.
\newblock \bibinfo{journal}{Photogrammetrie - Fernerkundung - Geoinformation}
  \bibinfo{volume}{2012}, \bibinfo{pages}{129--139}.
\newblock \URLprefix \url{http://dx.doi.org/10.1127/1432-8364/2012/0107},
  \DOIprefix\doi{10.1127/1432-8364/2012/0107}.
\bibitem[{Wei et~al.(2021)Wei, Zhang, Liu, Li and Li}]{WEI2021}
\bibinfo{author}{Wei, D.}, \bibinfo{author}{Zhang, Y.}, \bibinfo{author}{Liu,
  X.}, \bibinfo{author}{Li, C.}, \bibinfo{author}{Li, Z.},
  \bibinfo{year}{2021}.
\newblock \bibinfo{title}{Robust line segment matching across views via ranking
  the line-point graph}.
\newblock \bibinfo{journal}{ISPRS Journal of Photogrammetry and Remote Sensing}
  \bibinfo{volume}{171}, \bibinfo{pages}{49 -- 62}.
\newblock \URLprefix
  \url{http://www.sciencedirect.com/science/article/pii/S0924271620303014},
  \DOIprefix\doi{https://doi.org/10.1016/j.isprsjprs.2020.11.002}.
\bibitem[{{Wei} et~al.(2020){Wei}, {Zhang} and {Ji}}]{Wei2020}
\bibinfo{author}{{Wei}, Y.}, \bibinfo{author}{{Zhang}, K.},
  \bibinfo{author}{{Ji}, S.}, \bibinfo{year}{2020}.
\newblock \bibinfo{title}{Simultaneous road surface and centerline extraction
  from large-scale remote sensing images using cnn-based segmentation and
  tracing}.
\newblock \bibinfo{journal}{IEEE Transactions on Geoscience and Remote Sensing}
  \bibinfo{volume}{58}, \bibinfo{pages}{8919--8931}.
\newblock \DOIprefix\doi{10.1109/TGRS.2020.2991733}.
\bibitem[{Wei et~al.(2020)Wei, Jia, Jia, Khandelwal and Kumar}]{Wei2020b}
\bibinfo{author}{Wei, Z.}, \bibinfo{author}{Jia, K.}, \bibinfo{author}{Jia,
  X.}, \bibinfo{author}{Khandelwal, A.}, \bibinfo{author}{Kumar, V.},
  \bibinfo{year}{2020}.
\newblock \bibinfo{title}{Global river monitoring using semantic fusion
  networks}.
\newblock \bibinfo{journal}{Water} \bibinfo{volume}{12}.
\newblock \URLprefix \url{https://www.mdpi.com/2073-4441/12/8/2258},
  \DOIprefix\doi{10.3390/w12082258}.
\bibitem[{Weld et~al.(2019)Weld, Jang, Li, Zeng, Heimerl and
  Froehlich}]{Weld2019}
\bibinfo{author}{Weld, G.}, \bibinfo{author}{Jang, E.}, \bibinfo{author}{Li,
  A.}, \bibinfo{author}{Zeng, A.}, \bibinfo{author}{Heimerl, K.},
  \bibinfo{author}{Froehlich, J.E.}, \bibinfo{year}{2019}.
\newblock \bibinfo{title}{{Deep learning for automatically detecting sidewalk
  accessibility problems using streetscape imagery}}.
\newblock \bibinfo{journal}{ASSETS 2019 - 21st International ACM SIGACCESS
  Conference on Computers and Accessibility} ,
  \bibinfo{pages}{196--209}\DOIprefix\doi{10.1145/3308561.3353798}.
\bibitem[{Weng et~al.(2020)Weng, Xu, Xia, Zhang, Liu and Xu}]{Weng2020}
\bibinfo{author}{Weng, L.}, \bibinfo{author}{Xu, Y.}, \bibinfo{author}{Xia,
  M.}, \bibinfo{author}{Zhang, Y.}, \bibinfo{author}{Liu, J.},
  \bibinfo{author}{Xu, Y.}, \bibinfo{year}{2020}.
\newblock \bibinfo{title}{Water areas segmentation from remote sensing images
  using a separable residual segnet network}.
\newblock \bibinfo{journal}{ISPRS International Journal of Geo-Information}
  \bibinfo{volume}{9}.
\newblock \URLprefix \url{https://www.mdpi.com/2220-9964/9/4/256},
  \DOIprefix\doi{10.3390/ijgi9040256}.
\bibitem[{Xia et~al.(2019)Xia, Qian, Zhang, Liu and Xu}]{Xia2019}
\bibinfo{author}{Xia, M.}, \bibinfo{author}{Qian, J.}, \bibinfo{author}{Zhang,
  X.}, \bibinfo{author}{Liu, J.}, \bibinfo{author}{Xu, Y.},
  \bibinfo{year}{2019}.
\newblock \bibinfo{title}{{River segmentation based on separable attention
  residual network}}.
\newblock \bibinfo{journal}{Journal of Applied Remote Sensing}
  \bibinfo{volume}{14}, \bibinfo{pages}{1 -- 15}.
\newblock \URLprefix \url{https://doi.org/10.1117/1.JRS.14.032602},
  \DOIprefix\doi{10.1117/1.JRS.14.032602}.
\bibitem[{Yan et~al.(2019)Yan, Ai, Yang and Yin}]{Yan2019}
\bibinfo{author}{Yan, X.}, \bibinfo{author}{Ai, T.}, \bibinfo{author}{Yang,
  M.}, \bibinfo{author}{Yin, H.}, \bibinfo{year}{2019}.
\newblock \bibinfo{title}{{A graph convolutional neural network for
  classification of building patterns using spatial vector data}}.
\newblock \bibinfo{journal}{ISPRS Journal of Photogrammetry and Remote Sensing}
  \bibinfo{volume}{150}, \bibinfo{pages}{259--273}.
\newblock \URLprefix \url{https://doi.org/10.1016/j.isprsjprs.2019.02.010},
  \DOIprefix\doi{10.1016/j.isprsjprs.2019.02.010}.
\bibitem[{Yang and Chen(2015)}]{YANG2015}
\bibinfo{author}{Yang, B.}, \bibinfo{author}{Chen, C.}, \bibinfo{year}{2015}.
\newblock \bibinfo{title}{Automatic registration of uav-borne sequent images
  and lidar data}.
\newblock \bibinfo{journal}{ISPRS Journal of Photogrammetry and Remote Sensing}
  \bibinfo{volume}{101}, \bibinfo{pages}{262 -- 274}.
\newblock \URLprefix
  \url{http://www.sciencedirect.com/science/article/pii/S0924271615000180},
  \DOIprefix\doi{https://doi.org/10.1016/j.isprsjprs.2014.12.025}.
\bibitem[{{Yang} et~al.(2019){Yang}, {Li}, {Ye}, {Lau}, {Zhang} and
  {Huang}}]{Yang2019}
\bibinfo{author}{{Yang}, X.}, \bibinfo{author}{{Li}, X.},
  \bibinfo{author}{{Ye}, Y.}, \bibinfo{author}{{Lau}, R.Y.K.},
  \bibinfo{author}{{Zhang}, X.}, \bibinfo{author}{{Huang}, X.},
  \bibinfo{year}{2019}.
\newblock \bibinfo{title}{Road detection and centerline extraction via deep
  recurrent convolutional neural network u-net}.
\newblock \bibinfo{journal}{IEEE Transactions on Geoscience and Remote Sensing}
  \bibinfo{volume}{57}, \bibinfo{pages}{7209--7220}.
\newblock \DOIprefix\doi{10.1109/TGRS.2019.2912301}.
\bibitem[{Yavari et~al.(2018)Yavari, {Valadan Zoej} and Salehi}]{YAVARI2018}
\bibinfo{author}{Yavari, S.}, \bibinfo{author}{{Valadan Zoej}, M.J.},
  \bibinfo{author}{Salehi, B.}, \bibinfo{year}{2018}.
\newblock \bibinfo{title}{An automatic optimum number of well-distributed
  ground control lines selection procedure based on genetic algorithm}.
\newblock \bibinfo{journal}{ISPRS Journal of Photogrammetry and Remote Sensing}
  \bibinfo{volume}{139}, \bibinfo{pages}{46 -- 56}.
\newblock \URLprefix
  \url{http://www.sciencedirect.com/science/article/pii/S0924271618300595},
  \DOIprefix\doi{https://doi.org/10.1016/j.isprsjprs.2018.03.002}.
\bibitem[{{Zhang} et~al.(2020){Zhang}, {Cui} and {Zhu}}]{Zhang2020}
\bibinfo{author}{{Zhang}, Z.}, \bibinfo{author}{{Cui}, P.},
  \bibinfo{author}{{Zhu}, W.}, \bibinfo{year}{2020}.
\newblock \bibinfo{title}{Deep learning on graphs: A survey}.
\newblock \bibinfo{journal}{IEEE Transactions on Knowledge and Data
  Engineering} ,
  \bibinfo{pages}{1--1}\DOIprefix\doi{10.1109/TKDE.2020.2981333}.
\bibitem[{{Zhang} et~al.(2019){Zhang}, {Li}, {Bi}, {Zheng}, {Wang}, {Huang},
  {Luo}, {Xu} and {Gao}}]{Zhang_2019_CVPR}
\bibinfo{author}{{Zhang}, Z.}, \bibinfo{author}{{Li}, Z.},
  \bibinfo{author}{{Bi}, N.}, \bibinfo{author}{{Zheng}, J.},
  \bibinfo{author}{{Wang}, J.}, \bibinfo{author}{{Huang}, K.},
  \bibinfo{author}{{Luo}, W.}, \bibinfo{author}{{Xu}, Y.},
  \bibinfo{author}{{Gao}, S.}, \bibinfo{year}{2019}.
\newblock \bibinfo{title}{Ppgnet: Learning point-pair graph for line segment
  detection}, in: \bibinfo{booktitle}{2019 IEEE/CVF Conference on Computer
  Vision and Pattern Recognition (CVPR)}, pp. \bibinfo{pages}{7098--7107}.
\newblock \DOIprefix\doi{10.1109/CVPR.2019.00727}.

\end{thebibliography}

\end{document}